\documentclass[11pt]{article}

\usepackage[final]{acl}

\usepackage{times}
\usepackage{latexsym}

\usepackage[T1]{fontenc}

\usepackage[utf8]{inputenc}

\usepackage{microtype}

\usepackage{inconsolata}

\usepackage{graphicx}
\usepackage[utf8]{inputenc} 
\usepackage[T1]{fontenc}    
\usepackage{url}            
\usepackage{booktabs}       
\usepackage{amsfonts}       
\usepackage{nicefrac}       
\usepackage{microtype}      
\usepackage{xcolor}         
\usepackage{marvosym}
\usepackage{multirow}
\usepackage{makecell}
\usepackage{colortbl}
\usepackage{tcolorbox}
\usepackage{longtable}
\usepackage{cleveref}
\usepackage{subcaption}
\usepackage{pifont}
\usepackage{framed}
\usepackage[utf8]{inputenc}
\usepackage{longtable}
\definecolor{mypink}{rgb}{.99,.91,.95}
\definecolor{mygreen}{rgb}{.9,.99,.9}
\definecolor{myorange}{RGB}{255, 217, 192}
\definecolor{mygray}{gray}{.9}
\definecolor{shadecolor}{rgb}{0.9,0.9,0.9}
\title{Leave No Document Behind:\\ Benchmarking Long-Context LLMs with Extended Multi-Doc QA}

\author{%
  Minzheng Wang$^{1,3}$\thanks{Equal contribution.}, 
  Longze Chen$^{2,3}$\footnotemark[1],
  Cheng Fu$^{4}$,
  Shengyi Liao$^{4}$,
  Xinghua Zhang$^{4}$\\
  {\bf Bingli Wu}$^{4}$,
  {\bf Haiyang Yu}$^{4}$,
  {\bf Nan Xu}$^{1}$,
  {\bf Lei Zhang}$^{2,3}$,
  {\bf Run Luo}$^{2,3}$,
  {\bf Yunshui Li}$^{2,3}$\\
  {\bf Min Yang}$^{2}$\thanks{Corresponding author.}, 
  {\bf Fei Huang}$^{4}$, 
  {\bf Yongbin Li}$^{4}$\footnotemark[2]\\
  $^{1}$ MAIS, Institute of Automation, Chinese Academy of Sciences\\
  $^{2}$ Shenzhen Institute of Advanced Technology, Chinese Academy of Sciences\\
  $^{3}$ School of Artificial Intelligence, University of Chinese Academy of Sciences\\
  $^{4}$ Alibaba Group\\
  \small{
\Letter:~\texttt{wangminzheng2023@ia.ac.cn};~~\texttt{lz.chen2@siat.ac.cn}}
}

\definecolor{kellygreen}{rgb}{0.3, 0.73, 0.09}
\definecolor{alizarin}{rgb}{0.82, 0.1, 0.26}
\newcommand{\cmark}{{\color{kellygreen} \ding{51}}}
\newcommand{\xmark}{{\color{alizarin} \ding{55}}}

\begin{document}
\maketitle

\begin{abstract}
Long-context modeling capabilities have garnered widespread attention, leading to the emergence of Large Language Models (LLMs) with ultra-context windows. Meanwhile, benchmarks for evaluating long-context LLMs are gradually catching up.
However, existing benchmarks employ irrelevant noise texts to artificially extend the length of test cases, diverging from the real-world scenarios of long-context applications.
To bridge this gap, we propose a novel long-context benchmark, \textbf{Loong}, aligning with realistic scenarios through extended multi-document question answering (QA). Unlike typical document QA, in Loong's test cases, each document is relevant to the final answer, ignoring any document will lead to the failure of the answer.
Furthermore, Loong introduces four types of tasks with a range of context lengths: \textit{Spotlight Locating}, \textit{Comparison}, \textit{Clustering}, and \textit{Chain of Reasoning}, to facilitate a more realistic and comprehensive evaluation of long-context understanding.
Extensive experiments indicate that existing long-context language models still exhibit considerable potential for enhancement. Retrieval augmented generation (RAG) achieves poor performance, demonstrating that Loong can reliably assess the model's long-context modeling capabilities.\footnote{The code and benchmark are available at \url{https://github.com/MozerWang/Loong} and \url{https://github.com/AlibabaResearch/DAMO-ConvAI/tree/main/Loong}}

\end{abstract}
\begin{figure}[t]
    \centering
    \includegraphics[width=1.0\columnwidth]{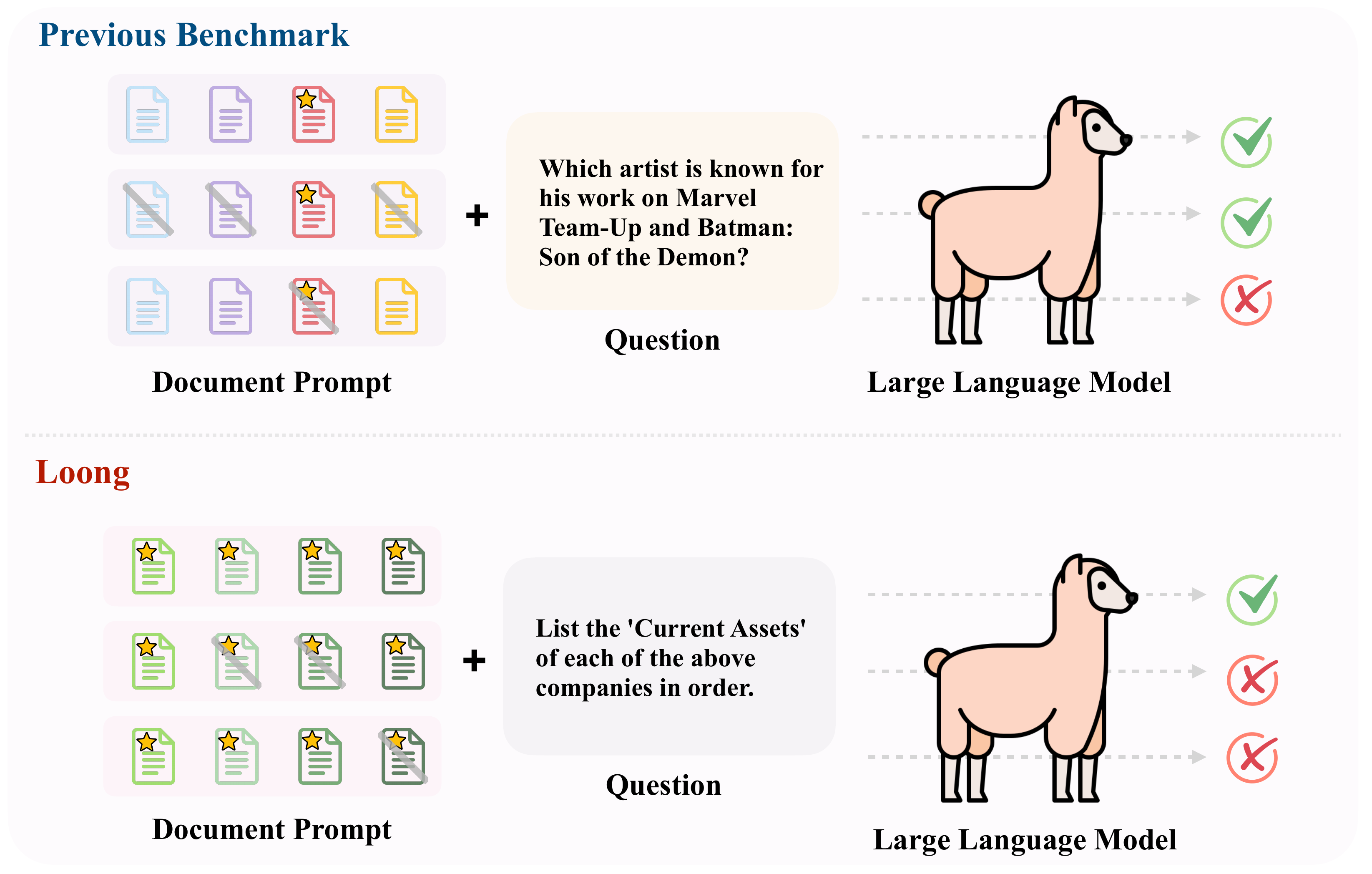}
    \caption{Previous benchmarks vs. Loong. \includegraphics[scale=0.015]{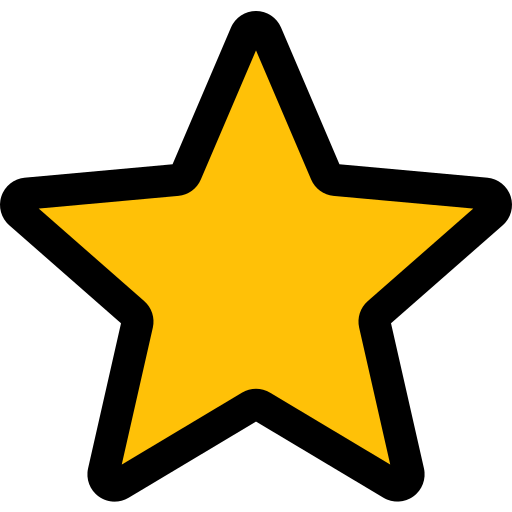} marks the existence of evidence related to the answer in that document. Compared to {\it \textbf{centralized}} distribution in previous ones, evidence in Loong are {\it \textbf{scattered}} in different parts across multi-document long contexts, necessitating that no document can be ignored for success.}
    \label{fig:overview}
\end{figure}

\section{Introduction}

Large Language Models (LLMs) have exhibited remarkable proficiency in diverse downstream applications~\cite{openai2024gpt4}. Recent works focus on scaling up the context window of LLMs~\citep{xiong2023effective, peng2023yarn,chen2024long}, which is crucial for LLMs in handling complex tasks that require delving deeply into long texts. A few LLM (e.g. GPT-4o, Gemini-Pro) websites have been equipped with the intelligent document analysis function, allowing users to upload documents for answering queries. These demand the model leverage its long-context capability to conduct an in-depth analysis of multiple long documents. Meanwhile, retrieval-augmented generation (RAG) has emerged as a widely adopted framework that prompts LLMs with multiple relevant retrieved contents and can significantly improve model performance~\cite{wu2024faithful, Chen_Lin_Han_Sun_2024}.

\begin{table*}[t]
\centering
{\setlength{\tabcolsep}{0.6pt}
\footnotesize
\scalebox{1.0}{\begin{tabular}{@{}l | c c c c c c @{}}
\toprule
Benchmark &  
\begin{tabular}[c]{@{}c@{}}\hspace{0.5em} Multi-doc \hspace{0.5em} \\ Tasks\end{tabular} & 
\begin{tabular}[c]{@{}c@{}}\hspace{0.5em} Broad \hspace{0.5em} \\ Length Sets\end{tabular} & 
\begin{tabular}[c]{@{}c@{}}\hspace{0.5em} Avoidance of \hspace{0.5em} \\ Contamination\end{tabular} & 
\begin{tabular}[c]{@{}c@{}}\hspace{0.5em} Realistic \hspace{0.5em} \\ Scenarios\end{tabular} & 
\begin{tabular}[c]{@{}c@{}}\hspace{0.5em} High Evidence \hspace{0.5em} \\ Dispersion\end{tabular} & 
\hspace{0.5em} Multilingual \hspace{0.5em} \\
\midrule
{\bf{L-Eval}}~\citep{an2023eval} & \xmark & \xmark & \xmark & \cmark & \xmark & \cmark \\
{\bf{LongBench}}~\citep{bai2023longbench} & \cmark & \xmark & \xmark & \cmark & \xmark & \cmark \\
{\bf{Marathon}}~\citep{zhang2023marathon} & \cmark & \cmark & \xmark & \cmark & \xmark & \cmark \\
{\bf{LooGLE}}~\citep{li2023loogle} & \xmark & \xmark & \xmark & \cmark & \xmark & \cmark \\
{\bf{InfiniteBench}}~\citep{zhang2024infty} & \cmark & \xmark & \xmark & \cmark & \xmark & \cmark \\
\midrule
{\bf{RULER}}~\citep{hsieh2024ruler} & \cmark & \cmark & \xmark & \xmark & \xmark & \xmark \\
{\bf{NIAH}}~\citep{needleinhaystack} & \xmark & \cmark & \cmark & \xmark & \xmark & \xmark \\
\midrule
\bf{Loong (Ours)} & \cmark & \cmark & \cmark & \cmark & \cmark & \cmark \\
\bottomrule
\end{tabular}}}
\caption{Characteristics of Loong, where the evidences are scattered across multi-document long contexts.}
\label{tab:related_benchmarks}
\end{table*}
However, there remains a lack of appropriate benchmarks for evaluating long-context understanding in real-world multi-document scenarios. Multi-document input as long-context modeling possesses extensive application scenarios of LLMs, such as analysis of financial reports over the years.
Nevertheless, most existing benchmarks only place emphasis on single-document long contexts~\citep{an2023eval,li2023loogle,needleinhaystack} or involve multi-document question answering settings by adding distracting information to the input of existing short-context QA datasets~\citep{hsieh2024ruler}.
As shown in \Cref{fig:overview}, evidence supporting the answer in previous benchmarks is relatively centralized, such as being contained within a single document.
Yet, such a centralized distribution of evidence may cause the model to overlook certain documents and take shortcuts to formulate an answer, simplifying the modeling of the real context.
Moreover, the prevalent evaluation tasks, such as ``\textit{needle in a haystack}'' (NIAH)~\citep{needleinhaystack}, only scratch the surface of long-context understanding by searching from context, far from real-world demands.

We commence with ``$\mathtt{leave}$ $\mathtt{no}$ $\mathtt{document}$ $\mathtt{behind}$'' and scatter the evidence across multi-document long contexts. In this context, bypassing any document will lead to an erroneous answer, which better tests the long-context modeling ability. To this end, this paper develops Loong, an innovative benchmark crafted to evaluate the long-context ability of LLMs across multiple documents in real-world scenarios.
Loong typically consists of 11 documents per test instance on average, nearly all of which are selected from the year 2024, spanning three real-world scenarios in English and Chinese: (1) \textit{Financial Reports}, (2) \textit{Legal Cases}, and (3) \textit{Academic Papers}. Meanwhile, Loong introduces new evaluation tasks from the perspectives of \textit{Spotlight Locating}, \textit{Comparison}, \textit{Clustering}, and \textit{Chain of Reasoning}, every test case is newly annotated. Furthermore, Loong features inputs of varying lengths (e.g., 10K-50K, 50K-100K, 100K-200K, \textgreater 200K) and evaluation tasks of diverse difficulty, enabling fine-grained assessment of LLMs across different context lengths and task complexities.

We conduct extensive experiments on Loong to test the long-context modeling capabilities of several advanced LLMs. The empirical results show that even the current most powerful LLMs still struggle with the tasks in Loong, suggesting significant room for improvement in current LLMs. Furthermore, this paper conducts in-depth analyses regarding the behavior of long-context LLMs involving RAG and the scaling law of context size.
Our main contributions are summarized as follows:
\begin{itemize}
\item Loong primarily focuses on testing the long-context modeling ability of LLMs across multiple documents by scattering the evidence to examine the actual length of long contexts.
\item Loong offers evaluation sets with varying input lengths and evaluation tasks of differing difficulty, encompassing novel task categories. All test instances are newly annotated and checked to guarantee quality.
\item Extensive experiments and analyses deeply unveil the performance of Long-Context LLMs, also offering insights into improving Long-Context modeling abilities.

\end{itemize}

\begin{figure*}[tp]
    \centering
    \includegraphics[width=1\textwidth]{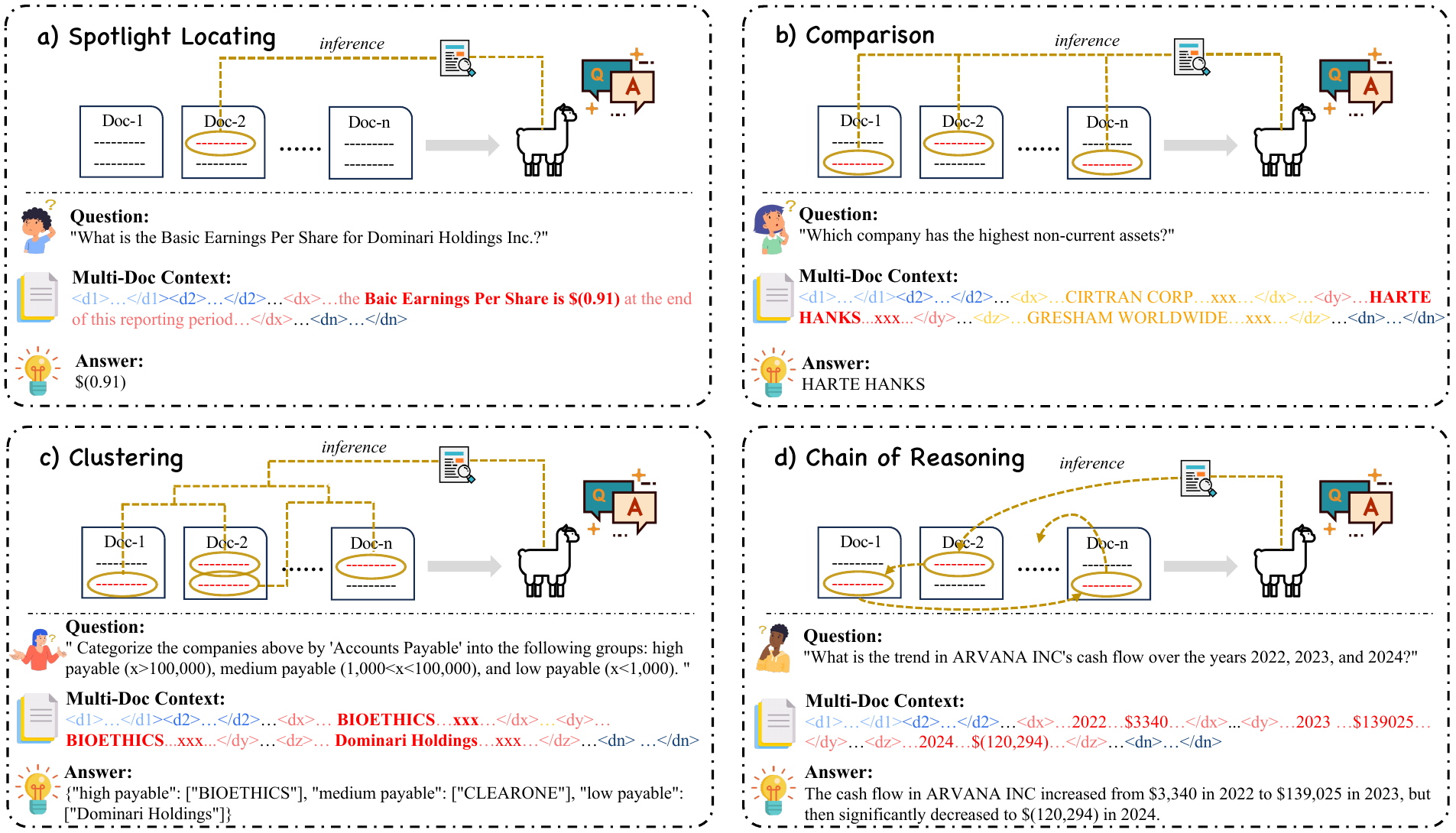} 
    \caption{Showcase of four evaluation tasks in Loong (<$\mathtt{di}$>...</$\mathtt{di}$> marks the content of the i-th document). a) \textit{Spotlight Locating}: Locate the evidence. b) \textit{Comparison}: Locate and compare the evidence. c) \textit{Clustering}: Locate and cluster the evidence into groups. d) \textit{Chain of Reasoning}: Locate and reasoning along a logical
chain.}
    \label{fig:main_fig}
\end{figure*}

\section{Related Work}
\subsection{Long-Context Language Models}
With support for increasingly larger context windows, closed-source LLMs have taken the lead in the field of long-context modeling. From 128k to 1000k, GPT-4o~\citep{openai2024gpt4}, Claude3-200k~\citep{anthropic2024claude} and Gemini-pro1.5-1000k~\citep{reid2024gemini} are capable of modeling increasingly longer documents, expanding the new scenarios that LLMs can handle.

Considering the quadratic complexity of Transformer~\citep{vaswani2017attention}, training LLMs with extensive context windows from scratch necessitates substantial computational resources, exceeding the capabilities of the general researchers.
Consequently, recent studies have explored ways to expand the context length of these models during the fine-tuning stage. For example, PI~\citep{chen2023extending}, NTK-aware~\citep{blocntkaware}, YaRN~\citep{peng2023yarn}, Giraffe~\citep{pal2023giraffe}, Code LLaMA~\citep{roziere2023code}, and PoSE~\citep{zhu2023pose} adapts position embedding based on the rotary position encoding (RoPE)~\citep{su2024roformer}, 
with only a few fine-tuning steps, the context length can be efficiently extended. 

Another strong baseline for long-context modeling is the sliding window method. Various sliding window-based variants such as ALibi~\citep{press2021train}, xPos~\citep{sun2022length}, PCW~\citep{ratner2022parallel}, LM-Infinit~\citep{han2023lm}, StreamingLLM~\citep{xiao2023efficient} are used to achieve efficient context scaling. Yet they diverge from the global perception characteristic of the Transformer, failing to exploit the entire context.

\subsection{Long-Context Benchmarks}
Long-context modeling methods are rapidly evolving, yet the quality of existing benchmarks does not align with this progress. 
Synthetic task such as Needle-in-a-Haystack (NIAH)~\citep{needleinhaystack} and Counting stars~\citep{song2024counting} are initially utilized for evaluating long-context language models (LCLMs) due to their lower construction costs, but they are indicative of only a surface form of long-context understanding.

Longbench~\citep{bai2023longbench}, LooGLE ~\citep{li2023loogle} and Marathon~\citep{zhang2023marathon} are earlier benchmarks for comprehensive assessment of long context. However, the average length for most tasks is between 5k and 25k, far less than the window size of LCLMs.
L-Eval~\citep{an2023eval}, BAMBOO~\citep{dong2023bamboo}, CLongEval~\citep{qiu2024clongeval} and InfiniteBench~\citep{zhang2024infty} contain sufficiently long evaluation data, and the wide variety of tasks makes the assessment more comprehensive.
RULER~\citep{hsieh2024ruler} creates a comprehensive testing method with flexibly adjustable length and difficulty, yet they only add distracting information to the input of existing short-context QA datasets.

While these long-context benchmarks have their advantages, there still lack a benchmark that is sufficiently long, free from data contamination~\citep{golchin2023time}, and fully aligned with the real-world multi-document question answering scenario. We conduct a detailed comparison with existing works in \Cref{tab:related_benchmarks}.

\subsection{Retrieval Augmented Language Models}
Leveraging long documents as external knowledge, Retrieval Augmented Language Models (RALMs) have achieved comparable or even better performance than LCLMs fine-tuned for specific tasks with long documents.
In previous study, RALMs could directly utilize the content retrieved during the inference phase. 
REPLUG~\citep{shi2023replug} treats the language model as a black box and the retrieval component as an adjustable plug-and-play module. 
RETRO~\citep{borgeaud2022improving} use a
chunked cross-attention module to incorporate the retrieved text. 
Additionally, \citet{xu2023retrieval} explored whether RALMs or LCLMs are more suitable for long-context tasks under a larger parameter setting. However, there is currently a lack of analysis on what tasks RALMs and LCLMs each excel at, thus making it difficult to determine which type a black box model belongs to.
\section{Loong: A Long-Context Benchmark}
\subsection{Overview}
The Loong benchmark comprises tasks across four categories: \textit{Spotlight Locating}, \textit{Comparison}, \textit{Clustering}, and \textit{Chain of reasoning}. To align with realistic scenarios, we collect documents from three domains: \textit{financial reports}, \textit{academic papers}, and \textit{legal cases}. Furthermore, all tasks are presented in the question-answering format, which are all newly annotated by GPT-4o and humans. Totally, Loong includes 1600 test instances in both Chinese and English, featuring four sets with different intervals of context size: $\mathtt{Set1}$ (10-50K), $\mathtt{Set2}$ (50-100K), $\mathtt{Set3}$ (100-200K) and $\mathtt{Set4}$ (200-250K). We use \texttt{tiktoken}\footnote{\url{https://platform.openai.com/tokenizer}} tokenizer to tokenize the input and report the number of tokens. \Cref{tb:stat} and \Cref{paper:length} show the details of data statistics. \Cref{paper:evidence_distribution} presents a comparison of evidence distribution between Loong and LongBench. The following sections will provide a detailed description of the evaluation task and benchmark construction.

\subsection{Evaluation Task}
\begin{table}
  \centering
  \resizebox{1\columnwidth}{!}{
    \begin{tabular}{l|c|c|c}
\toprule
\textbf{Category}&\textbf{Avg Token}&\textbf{Language}&\textbf{\#Test Instance}\\
\midrule
\rowcolor{mygray}\multicolumn{4}{c}{\textit{Task}} \\
\midrule
Spotlight Locating & 119.3K & EN, ZH & 250 \\
Comparison & 110.6K & EN, ZH & 300 \\
Clustering & 109.8K & EN, ZH & 641 \\
Chain of Reasoning & 103.9K & EN, ZH & 409 \\
\midrule
\rowcolor{mygray}\multicolumn{4}{c}{\textit{Sub Task}} \\
\midrule
Sequential Enumeration & 103K & EN, ZH & 87 \\
Extremum Acquisition & 115K & EN, ZH & 143 \\
Range Awareness& 111K & EN, ZH & 70 \\
Report Integration&  117K & EN, ZH & 250 \\
Citation\&Reference & 105K & EN & 270 \\
Case Classification & 106K & ZH & 121 \\
Temporal Analysis & 112K & EN, ZH & 100 \\
Citation Chain & 91K & EN & 130 \\
Link the Links & 117K & ZH & 113 \\
Solitaire & 94K & ZH & 66 \\
\midrule
\rowcolor{mygray}\multicolumn{4}{c}{\textit{Domain}} \\
\midrule
Financial Reports & 117.5K & EN, ZH & 700 \\
Legal Cases & 107.2K & ZH & 500 \\
Academic Papers & 100.9K & EN & 400 \\

\midrule
\rowcolor{mygray}\multicolumn{4}{c}{\textit{Length Set}} \\
\midrule
$\mathtt{Set1}$ (10-50K) & 37.8K & EN, ZH & 323 \\
$\mathtt{Set2}$ (50-100K) & 75.6K & EN, ZH & 564 \\
$\mathtt{Set3}$ (100-200K) & 138.9K & EN, ZH & 481 \\
$\mathtt{Set4}$ (200-250K) & 233.9K & EN, ZH & 232 \\
      \bottomrule
    \end{tabular}
  }
\caption{Data statistics of Loong benchmark.}
\label{tb:stat}
\end{table}
Based on various multi-document semantic relationships and LLMs' handling of multi-document input, we propose new task categories for multi-document long-context modeling and closer alignment with real-world scenarios. \Cref{fig:main_fig} illustrates the evaluation tasks of the Loong benchmark. \Cref{paper:Case} shows the detailed test case and prompt of each task.

\subsubsection{Spotlight Locating} 
The spotlight locating task is designed to assess the model's capability for knowledge localization, which constitutes the foundation ability of long-context processing. 
In this task, the evidence is contained in only one of multiple documents, which is the atomic setting of the key information locating.
The spotlight locating task is aimed at examining the LLMs' ability to search for evidence within one document from multiple ones. Other documents, which are in the same domain and have similar semantics as the document but are unrelated to the question, will serve as noise texts. The upper left of \Cref{fig:main_fig} provides an example of the spotlight locating task.

\subsubsection{Comparison} 
The comparison task is primarily aimed at evaluating the model's ability to compare multi-source information with long contexts. In this event, the evidence supporting the answer are distributed across multiple documents, testing the LLMs' ability to locate dispersed evidence and to correlate and compare them.

Comparison task includes three sub-tasks: 1) \textit{Sequential Enumeration}: Based on the concrete numerical value of a specific attribute, it requires the model to list all specific values corresponding to that attribute across multiple documents in a given order. 2) \textit{Extremum Acquisition}: It requires the model to deduce the extremum of all values corresponding to certain attributes in multiple documents. 3) \textit{Range Awareness}: Given a specific numerical or conceptual range, the model should output all objects within multiple documents that meet the condition.
The upper right of \Cref{fig:main_fig} gives an example of comparison task.
\subsubsection{Clustering} 
The clustering task entails an assessment of the model's ability to aggregate multi-source information from multi-document long contexts. The task requires the LLM to perform two main functions: first, to extract relevant data from multiple documents, and second, to integrate this information based on specific criteria. Ultimately, the LLM must cluster related evidence spread across these documents into coherent groups.

The clustering task encompasses three sub-tasks: 1) \textit{Report Integration}: This sub-task requires the model to group the evidence existing in the provided financial reports into corresponding sets based on textual or numerical criteria. 2) \textit{Citation\&Reference}: For a given paper, the model is tasked with identifying its citations and references from the candidate papers. 3) \textit{Case Classification}: Given the causes of several legal cases, the model is required to accurately categorize judgment documents.
The bottom left of \Cref{fig:main_fig} depicts an example of the clustering task.

\subsubsection{Chain of Reasoning} 
The chain of reasoning task requires the model to engage in multi-document reasoning along a logical pathway. This task evaluates the model's proficiency in logical reasoning, which requires LLMs to locate the corresponding evidence within multiple documents and model the logical relationships among them for deducing the answer.

The chain of reasoning task contains four sub-tasks: 
1) \textit{Temporal Analysis}: This task requires the model to analyze the changes or trends of a particular attribute based on the temporal relationship, such as taking into account the financial reports of a certain company over consecutive years or multiple quarters. 
2) \textit{Citation Chain}: This task requires the model to accurately understand each paper's content and its interconnections, ultimately inferring the linear citation relationships among them.
3) \textit{Link the Links}: This task involves presenting fact descriptions and trial results from different judgment documents separately. The model is tasked with accurately pairing each fact description with its corresponding trial result.
4) \textit{Solitaire}: This task first requires the model to match causes of action with judgment documents correctly, and then to sequentially infer multiple judgment documents based on the given sequence of causes of action.
The bottom right of \Cref{fig:main_fig} gives an example of the chain of reasoning task.

\begin{table*}[t]
\centering  
\resizebox{\textwidth}{!}{
\begin{tabular}{l|cc|cc|cc|cc|cc}
\toprule
\multirow{1}{*}{\textbf{Model}} & \multicolumn{2}{c|}{\textbf{Spotlight Locating}} & \multicolumn{2}{c|}{\textbf{Comparison}} & \multicolumn{2}{c|}{\textbf{Clustering}} & \multicolumn{2}{c|}{\textbf{Chain of Reasoning}} & \multicolumn{2}{c}{\textbf{Overall}}\\

\midrule
GPT-4o (128K) & 73.95 &0.62 & 50.50 &0.28 &44.29 &0.09 &57.95 &0.28 &53.47 &0.26 \\

\rowcolor{mygray}Gemini-1.5-pro (1000K) & 75.02 & 0.56 & 49.94 & 0.27 & 44.10 & 0.09 & 64.97 & 0.37 & 55.37 & 0.27 \\

Claude3.5-Sonnet (200K) & 58.45 & 0.49 & 54.21 & 0.35 & 45.77 & 0.07 & 43.92 & 0.25 & 48.85 & 0.23\\

\rowcolor{mygray}Qwen2-72B-Instruct (128K) & 54.17 & 0.36 &42.38 & 0.20 & 36.71 & 0.04 & 47.76 & 0.18 & 43.29 & 0.15\\ 

Claude3-Haiku (200K) & 68.68 & 0.59 & 42.10 & 0.21 & 35.04 & 0.02 & 47.59 & 0.17 & 44.88 & 0.19\\

\rowcolor{mygray}Kimi-Chat (200k) & 60.98 & 0.50 & 34.74 & 0.13 & 28.76 & 0.04 & 38.52 & 0.15 & 37.49 & 0.16 \\

GLM4-9B-Chat (1000K) & 57.35 & 0.47 & 40.38 & 0.20 & 28.52 & 0.02 & 39.94 & 0.16 & 38.31 & 0.16\\

\bottomrule
\end{tabular}
}
\caption{Overall results on four evaluation tasks. For each task, the indicator on the left represents the \textbf{\textit{Avg Scores} (0\textasciitilde100)}, while the right one represents the \textbf{\textit{Perfect Rate} (0\textasciitilde1)}.}
\label{tb:exp2}
\end{table*}
\begin{table*}[t]
\centering  
\resizebox{\textwidth}{!}{
\begin{tabular}{l|cc|cc|cc|cc|cc}
\toprule

\multirow{1}{*}{\textbf{Model}} & \multicolumn{2}{c|}{\textbf{Spotlight Locating}} & \multicolumn{2}{c|}{\textbf{Comparison}} & \multicolumn{2}{c|}{\textbf{Clustering}} & \multicolumn{2}{c|}{\textbf{Chain of Reasoning}} & \multicolumn{2}{c}{\textbf{Overall}}\\
\midrule
\multicolumn{11}{c}{\textbf{$\mathtt{Set1}$ (10K-50K)}} \\
GPT-4o (128K) & 85.67 &0.81 & 64.27 &0.33 &57.01 &0.24 &81.58 &0.55 &70.40 &0.44 \\

\rowcolor{mygray}Gemini-1.5-pro (1000K) & 75.00 & 0.60 & 54.88 & 0.28 & 56.15 & 0.23 & 70.64 & 0.37 & 63.36 & 0.34 \\

Claude3.5-Sonnet (200K) & 60.85 & 0.55 & 69.07 & 0.47 & 58.63 & 0.13 & 68.57 & 0.50 & 63.69 & 0.37\\

\rowcolor{mygray}Qwen2-72B-Instruct (128K) & 68.49 & 0.55 & 60.60 & 0.37 & 47.08 & 0.08 & 70.39 & 0.36 & 60.11 & 0.29\\ 

Claude3-Haiku (200K) & 60.94 & 0.55 & 59.97 & 0.40 & 45.53 & 0.04 & 66.85  & 0.34 & 57.14 & 0.28\\

\rowcolor{mygray}Kimi-Chat (200k) & 81.11 & 0.74 & 46.70 & 0.20 & 47,84 & 0.07 & 53.77 & 0.17 & 55.02 & 0.24 \\

GLM4-9B-Chat (1000K) & 63.11 & 0.53 & 54.10 & 0.27 & 39.50 & 0.08 &56.32  & 0.28 & 51.43 & 0.25\\

\midrule
\multicolumn{11}{c}{\textbf{$\mathtt{Set2}$ (50K-100K)}} \\
GPT-4o (128K) & 86.76 &0.72 & 59.81 &0.40 &47.83 &0.11 &62.09 &0.34 &58.38 &0.29 \\

\rowcolor{mygray}Gemini-1.5-pro (1000K) & 76.50 & 0.57 &  54.51 & 0.34 & 44.58 & 0.09 &  64.87& 0.34 & 55.56 & 0.26 \\

Claude3.5-Sonnet (200K) & 63.83 & 0.53 & 58.90 & 0.39 & 50.96 & 0.10 & 46.09 & 0.26 & 52.73 & 0.24\\

\rowcolor{mygray}Qwen2-72B-Instruct (128K) & 64.53 & 0.43 & 42.60 & 0.21 & 38.52 & 0.05 & 51.18 & 0.20 & 45.71 & 0.17\\ 

Claude3-Haiku (200K) & 73.71 & 0.66 & 41.90 & 0.22 & 36.18 & 0.02 & 50.20 & 0.15 & 45.45 & 0.17\\

\rowcolor{mygray}Kimi-Chat (200k) & 72.82 & 0.52 & 46.77 & 0.21 & 33.46 & 0.06 & 40.51 & 0.15 & 42.40 & 0.16 \\

GLM4-9B-Chat (1000K) & 65.04 & 0.54 & 41.80 & 0.23 & 30.72 & 0.02 & 42.34 & 0.17 & 40.19 & 0.17\\

\midrule
\multicolumn{11}{c}{\textbf{$\mathtt{Set3}$ (100K-200K)}} \\
GPT-4o (128K) & 74.84 &0.65 & 42.40 &0.21 &38.70 &0.04 &45.06 &0.09 &46.95 &0.19 \\

\rowcolor{mygray}Gemini-1.5-pro (1000K) & 81.25 & 0.56 & 44.66 & 0.20 & 39.90 & 0.05 & 58.38 & 0.36 & 52.05 & 0.24 \\

Claude3.5-Sonnet (200K) & 65.36 & 0.56 & 50.32 & 0.34 & 37.79 & 0.03 & 25.95 & 0.11 & 42.06 & 0.19\\

\rowcolor{mygray}Qwen2-72B-Instruct (128K) & 46.99 & 0.27 & 37.06 & 0.13 & 31.50 & 0.02 & 35.01 & 0.07 & 35.94 & 0.09\\ 

Claude3-Haiku (200K) & 77.81 & 0.67 & 37.07 & 0.17 & 30.94 & 0.01 & 36.87 & 0.12 & 41.41 & 0.18\\

\rowcolor{mygray}Kimi-Chat (200k) & 62.13 & 0.54 & 24.20 & 0.05 & 21.98 & 0.01 & 31.02 & 0.14 & 31.37 & 0.14 \\

GLM4-9B-Chat (1000K) & 69.19 & 0.56 & 37.99 & 0.18 & 26.63 & 0.01 & 32.30 & 0.09 & 37.36 & 0.16\\

\midrule
\multicolumn{11}{c}{\textbf{$\mathtt{Set4}$ (200K-250K)}} \\
GPT-4o (128K) & 36.79 &0.19 & 23.97 &0.08 &30.40 &0.00 &32.89 &0.07 &31.11 &0.07 \\

\rowcolor{mygray}Gemini-1.5-pro (1000K) & 62.23 & 0.49 & 43.08 & 0.20 & 36.48 & 0.00 & 68.51 & 0.49 & 50.70 & 0.25 \\

Claude3.5-Sonnet (200K) & 36.91 & 0.24 & 28.82 & 0.05 & 28.68 & 0.00 & 28.77 & 0.08 & 30.51 & 0.08\\

\rowcolor{mygray}Qwen2-72B-Instruct (128K) & 33.18 & 0.16 & 26.59 & 0.08 & 29.84 & 0.01 & 25.81 & 0.04 & 28.92 & 0.06\\ 

Claude3-Haiku (200K) & 53.26 & 0.40 & 27.00 & 0.03 & 25.36 & 0.00 & 28.11 & 0.05 & 32.15 & 0.10\\

\rowcolor{mygray}Kimi-Chat (200k) & 20.17 & 0.12 & 9.17 & 0.00 & 5.65 & 0.00 & 22.61 & 0.11 & 13.50 & 0.05 \\

GLM4-9B-Chat (1000K) & 15.67 & 0.12 & 21.33 & 0.05 & 12.35 & 0.00 & 21.04 & 0.05 & 16.84 & 0.05\\

\bottomrule
\end{tabular}
}
\caption{The performance of LLMs on four evaluation tasks with different length sets. For each task, the indicator on the left represents the \textbf{\textit{Avg Scores} (0\textasciitilde100)}, while the right one represents the \textbf{\textit{Perfect Rate} (0\textasciitilde1)}.}
\label{tb:exp1}
\end{table*}
\subsection{Benchmark Construction}
\subsubsection{Data Collection} 
We established six criteria for the manual collection of the required English and Chinese documents: 
(1) \textit{Timeliness}: The majority of the documents are the latest ones from the year 2024; 
(2) \textit{Accessibility}: The data is publicly available and permitted for download and collection; 
(3) \textit{Appropriate Length}: Collecting longer documents as much as possible and ensure they fit within the four designated length sets; 
(4) \textit{Parseability}: Chosen documents are easy to process and parse, facilitating conversion into natural language text; 
(5) \textit{Categorizability}: Documents can be manually sorted based on certain attributes, such as case type, research theme, or company category, allowing for organized archival; 
(6) \textit{Authoritativeness}: All documents are collected from scratch from official websites (e.g. China Judge Online, U.S. SEC, cninf, Arxiv, Semantic Scholar), ensuring the quality and authority of the documents. The detailed URL can be seen in \Cref{paper:doc_url}.

Specifically, regarding financial reports, we primarily collect the latest quarterly and annual reports for the year 2024, totaling 574 documents. 
For legal documents, our collection consists exclusively of cases adjudicated by the higher and intermediate courts in 2024, amounting to 629 documents. 
As for academic papers, our focus is on procuring the latest articles from arXiv in 2024, with a total of 764 papers. 
Additionally, to meet the requirements of the chain of reasoning task, we gather a small portion of financial reports and academic papers from before 2024. 
Upon the collection of documents, we first parse these documents, converting them uniformly into \texttt{TXT} format. Subsequently, we carry out further data cleansing, removing any portions that contain personal information.

\subsubsection{Annotation Process} 
Compared to annotating short texts, annotating long texts is more challenging. To address this issue, we designed innovative annotation workflows to reduce the cost of annotation while ensuring quality.

For \textit{financial reports}, we compress the information contained within the long context, breaking down the annotation process into numerous simple tasks. We initially manually identify hundreds of key attributes that cover the important information in the long context. Subsequently, we employ GPT-4o to execute the relatively simple task of information extraction, pulling the values corresponding to these key attributes. 
After obtaining the key attributes and their corresponding values, we can proceed to annotate only the compressed information, eliminating the need to refer back to the original lengthy texts. For \textit{legal cases}, we follow the classification provided by China Judge Online, manually downloading judgment documents sorted by different causes of action and case types. Additionally, we use a rule-based method to segment each judgment document into its factual statement and verdict sections. For \textit{academic papers}, we leverage the Semantic Scholar website's API to access the target paper's citations and references. Moreover, by utilizing the bbl files of each arXiv paper, we write scripts to recursively collect articles that meet the requirements of the linear citation chain task.

During the question-and-answer annotation phase, we adopt two approaches: (1) \textit{Template-based}: We design question types and templates, and based on pre-classified documents, we construct Q\&A pairs using rules. (2) \textit{Free annotation}: Referring to the compressed information of multiple documents, we design prompts with four different task descriptions. We employ GPT-4o to generate Q\&A pairs for each task.

\subsubsection{Quality Control}
Throughout the annotation process, we employ several methods to ensure accuracy: (1) \textit{Evidence Recall}: By designing prompts that not only prompt GPT-4o to generate labels but also to recall evidence supporting the labels from the text, significantly enhancing the accuracy in practical applications. (2) \textit{Self-Check}: GPT-4o reviews the original text to re-evaluate and correct any mistakes in the generated labels. (3) \textit{Manual Check}: We manually review and confirm the quality of annotations, eliminating any unreasonable or low-quality questions. Additionally, we also take into account the distribution and number of different length sets, sub-tasks, and language. From a pool of 2,814 entries, we conduct a secondary selection process, ultimately choosing 1,600 entries for our final benchmark.

\section{Experiments}
\subsection{Experimental Setup}
\textbf{Models} We evaluate six advanced long-context LLMs, with their context window sizes ranging from 128K to 1000K, including API-based LLMs: GPT-4o-128K~\citep{openai2024gpt4}, Gemini-1.5-pro-1000K~\citep{reid2024gemini}, Claude3.5-Sonnet-200K~\citep{anthropic2024claude35}, Claude3-Haiku-200K~\citep{anthropic2024claude}, Kimi-Chat-200K\footnote{\url{https://kimi.moonshot.cn/}} and Open-sourced LLMs: Qwen2-72B-Instruct-128K~\citep{bai2023qwen}, GLM4-9B-Chat-1000K~\citep{du2022glm}.

\noindent\textbf{Evaluation Metric} In the long-context question-answering scenarios, traditional evaluation metrics F1 and Rouge-L may lead to inaccurate responses.
Recent research~\citep{zhang2023wider, liu-etal-2024-calibrating-llm, wang2024evaluating} indicates that the GPT-4~\citep{openai2024gpt4} evaluator demonstrates high consistency with human evaluations, making it a reasonably reliable annotator. Building on these considerations, we prompt GPT-4 as a judge to evaluate the model's output based on the golden answer and the question's requirements from three aspects: \textit{Accuracy}, \textit{Hallucinations}, and \textit{Completeness}, scoring from 0 to 100. For a detailed prompt, please refer to the \Cref{paper:GPT4-as-the-Judge}. We also design two indicators: (1) \textit{Avg Scores}: the average value of scores given by GPT-4 for all questions; (2) \textit{Perfect Rate}: the proportion of cases scoring 100 out of the total cases. The latter is a more stringent evaluation metric compared to the former.

\noindent\textbf{Prompt Templates}
For different sub-tasks, we require the model to follow the given instructions and output the answer according to the specific prompts shown in \Cref{paper:Case}. 

\noindent\textbf{Input Truncation} 
Due to input length limits, we assess whether adding a document would exceed the model's processing length when concatenating multiple documents. If appending the document would surpass the model's capacity, we discard it from the concatenation process. The evaluation and selection process continues until we have reviewed all documents that need concatenation. 

\noindent\textbf{Implement Details}
We set `temperature = 0' to eliminate randomness and keep other hyper-parameters default. For API-Based LLMs, we directly utilize the official API for testing. Since the Kimi-Chat-200k currently does not provide an interface, we manually input content on the web. As for open-source models, we conduct experiments on a server with 8$\times$A100 80GB.

\subsection{Main Results}
We assess seven advanced LLMs on the Loong benchmark. The main results are shown in \Cref{tb:exp2} and \Cref{tb:exp1}. We can see that Gemini-1.5-pro shows the best overall performance, especially excelling in the processing of ultra-long context within $\mathtt{Set3}$ and $\mathtt{Set4}$. Its comprehensive score reached 55.37 with the perfect rate of 27\%, followed by GPT-4o. Besides, the long-context modeling capacity of open-source models still falls short when compared to that of the most powerful closed-source models in the Loong. Additionally, larger-parameter models outperform their smaller counterparts within the same window size, indicating the advantages of scaling up model sizes for improved long-context modeling. The overall assessment results highlight that even the most advanced long-context LLMs currently fail to achieve passing marks, particularly in terms of the perfect rate. This suggests that there exists significant room for improvement in the long-context modeling capabilities of LLMs. We also compare the current results with other current Long-Context benchmarks, which can be seen in \Cref{paper: comparison_benchmark}.
\subsection{Task Analysis}
Analyzing performance across different tasks, models exhibit their best performance in the spotlight locating task. This can be attributed to the task's relative simplicity, which tests the foundational capabilities of long-context modeling. Moreover, the evidence is only distributed within a single document, making it easier to locate and less prone to confusion. In contrast, due to the requirements of multi-source information inference, the comparison and cluster tasks present greater challenges, leading to model underperformance. These tasks necessitate not only the collection of evidence across documents but also involve complex reasoning processes such as matching, contrasting, and classification. Thus, they more rigorously test the higher-order capabilities of long-context modeling, revealing significant gaps in the current models' abilities. Regarding the chain of reasoning task, models perform well within Set1. However, as the context length increases, their performance drastically declines. This suggests that within the scope of long-context modeling capabilities, LLMs possess adequate skills in temporal analysis, logical sequencing, and linking multiple concepts. Nevertheless, an overflow in context length leads to the loss of key evidence, severely impacting the accuracy of chain reasoning tasks.
\begin{figure}[t]
    \centering
    \begin{subfigure}[b]{1\columnwidth}
        \centering
        \includegraphics[width=\columnwidth]{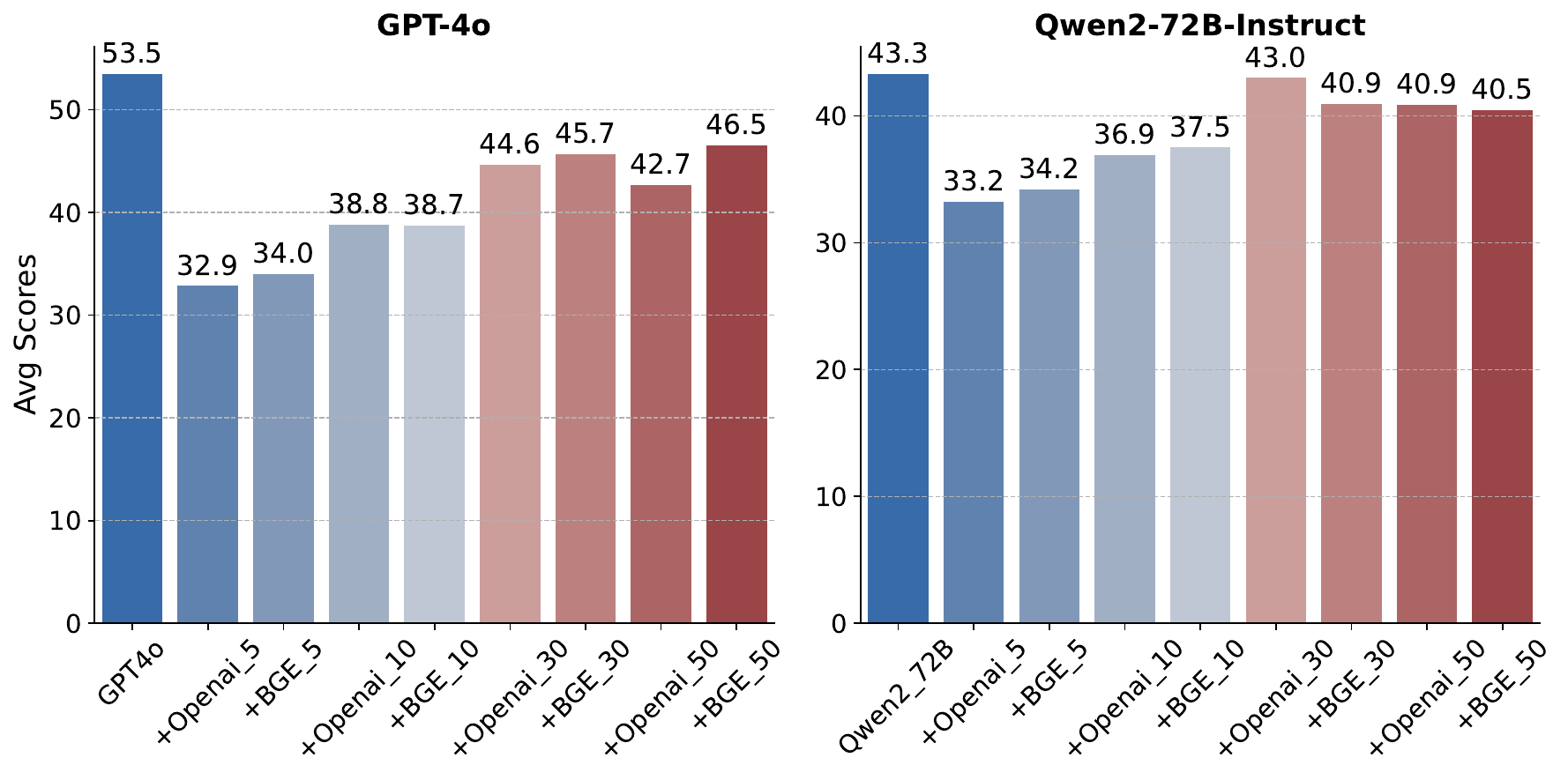}
        \caption{The overall results on all length sets.}
        \label{fig:rag1}
    \end{subfigure}
    \begin{subfigure}[b]{1\columnwidth}
        \centering
        \includegraphics[width=\columnwidth]{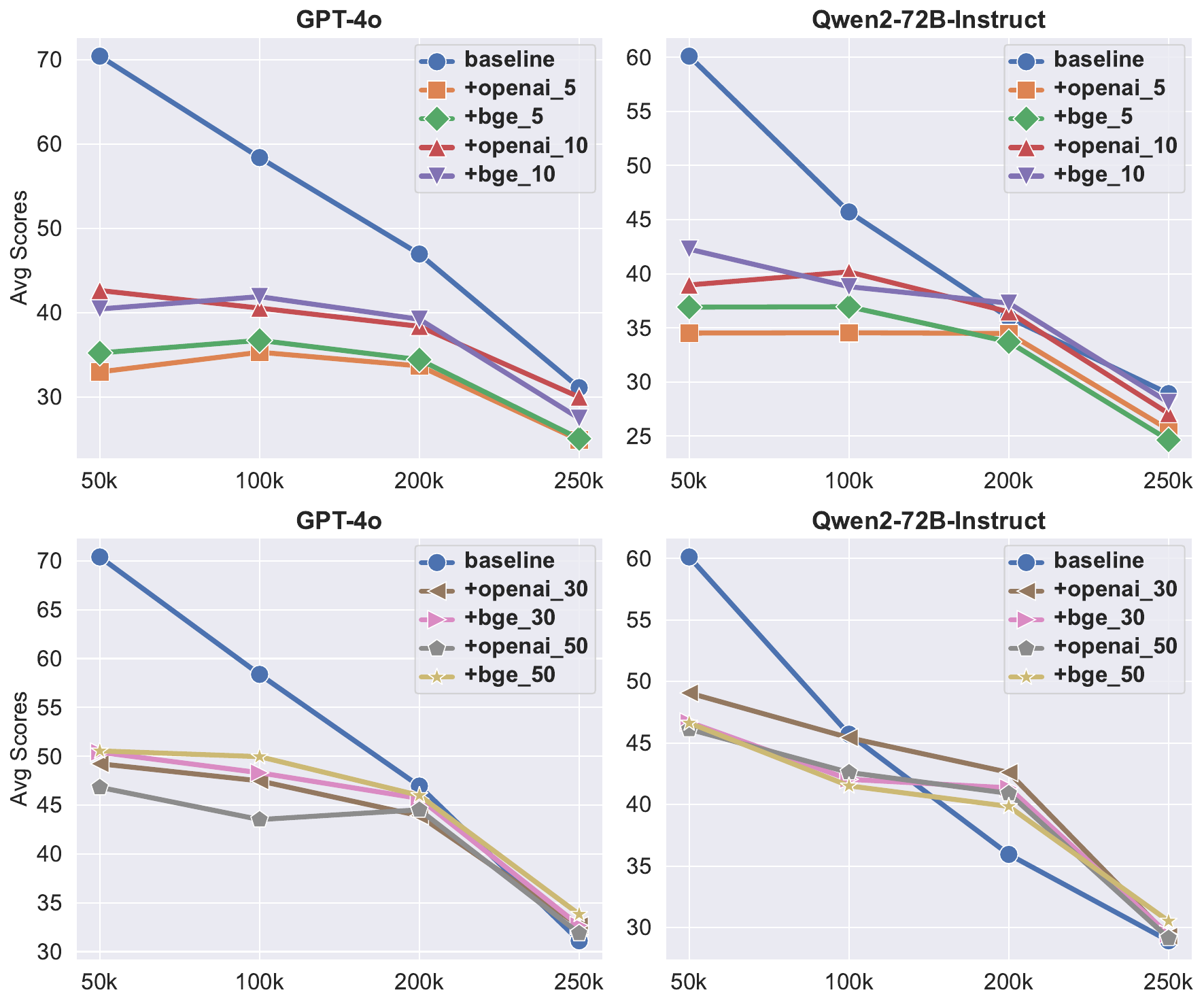}
        \caption{The detailed results on different length sets. The baseline means the setting without RAG.}
        \label{fig:rag2}
    \end{subfigure}
    \caption{The results on all tasks after adding RAG module. We only represents the \textbf{\textit{Avg Scores} (0\textasciitilde100)}.}
    \label{fig:rag}
\end{figure}

\subsection{Scaling Law of Context Window}
It's observed that the general performance of all models deteriorates with the increase in context size. 
As observed from \Cref{tb:exp1}, it is apparent that for the same task, models perform well within small length sets but exhibit a notable performance decline as the length increases. This indicates that the models possess a certain capability to process the task, yet their performance is constrained by the context window. Moreover, despite being trained on 128K data, the GPT-4o and Qwen2-72B-Instruct begin to show performance degradation within the 50-100K interval, revealing that their actual capability boundary is significantly lower than the claimed window size. This suggests the presence of an ineffective zone within the claimed window. There exists a Scaling Law for model window sizes: to truly equip an LLM with the ability to handle 128K long texts, it should be trained on data exceeding 128K, meaning the training length should be greater than the actual processable length. Among numerous models, only the Gemini is less affected by changes in context length, which was training on the ultra-long context of 1000K. To ensure your model genuinely possesses the desired context window size, train it on longer data!
\subsection{RAG or Not}
We have also incorporated the Embedding RAG module into the GPT-4o and Qwen2-72B-Instruct to explore whether RAG can enhance the model's performance on Loong. 
For the Embedding choice, we employ two distinct models: the OpenAI Embedding model\footnote{\url{https://huggingface.co/Xenova/text-embedding-ada-002}} and the BGE Embedding model\footnote{\url{https://huggingface.co/BAAI/bge-m3}}. Besides, we set the top-k value of 5, 10, 30, and 50 for each model respectively, and the chunk size is 1024. The result is shown in \Cref{fig:rag} and the details can be seen in \Cref{paper:rag}.

\noindent\textbf{Benchmark Analysis}
It is evident that the inclusion of RAG does not enhance the model's overall performance on the Loong, and there is a noticeable decline in assessment. This is because the evidence in the Loong is distributed relatively evenly across multiple documents, requiring a comprehensive understanding of long texts by the model. RAG, being more limited, only shows some effectiveness in the task with sparse evidence, such as spotlight locating. However, RAG's negative impact is significant for tasks requiring a high level of comprehensiveness. Integrating RAG does not enhance the performance, indicating that Loong focuses on evaluating the model's complex reasoning and comprehensive analysis capabilities for long contexts, thereby effectively assessing the LLM's long-context modeling ability.

\noindent\textbf{Model Analysis}
Comparing the performance between GPT-4o and Qwen2-72B-Instruct, it is evident that the powerful long-context LLM significantly outperforms RAG. On the other hand, the performance of RAG is closer to the original performance when used with a weak context model. This is because a strong long-context LLM can fully exploit the complete information flow of long contexts, capturing complex dependencies and semantic information. However, RAG causes context fragmentation and information loss, impairing the model's understanding and reasoning capabilities, thereby preventing the full utilization of its inherent modeling advantages. Consequently, a strong long-text modeling capability is not suitable for enhancement through RAG. Conversely, a weak model with poor long-context modeling capability cannot effectively capture information, and RAG cannot compensate for this deficiency.

\noindent\textbf{Length Analysis}
Within the context window size that the model can handle, RAG does not offer an advantage. However, for ultra-long context sets, a high top-k setting of RAG can produce certain effects. This is because, in short context sets, the model's inherent modeling capability can effectively handle the entire text length without losing information. The introduction of RAG, conversely, may result in the loss of certain evidence, leading to information gaps. In ultra-long context collections, RAG can effectively compress information, recalling evidence that the LLM could not access due to length truncation, thereby enhancing the model's performance on Loong.

We also provide analytical experiment to show that RAG does not cover all evidence, which can be seen in \Cref{paper:RAG_recall_rate}. Relying on RAG cannot resolve all the problems associated with long-text modeling. To genuinely improve the long-context modeling capability, stronger training methods and effective training on longer texts are required, rather than merely integrating the RAG module.
\section{Conclusion}
In this study, we propose Loong, a question-answering format benchmark designed to evaluate long-context comprehension in real-world multi-document scenarios. We analyze six advanced language models (LLMs), considering variations in their parameter sizes and context windows, including GPT-4o and Gemini-Pro1.5. Notably, even the most powerful long-context LLMs fail to achieve satisfactory performance. Furthermore, we conduct in-depth analyses to enhance long-context modeling capabilities by comparing the RAG approach and the scaling laws related to context size. 
\section*{Limitations}
Here we list some of the limitations that are not considered when designing Loong: (1) Limited Domains. The purpose of Loong is to evaluate the long-context understanding capabilities in real-world multi-document scenarios. However, a sea of multi-document domains exists in the real world.
Considering annotation costs and model evaluation efficiency, we only cover the most representative parts of them: financial, legal, and academic. (2) High Annotation Cost. To enhance the reliability of Loong in assessing the LLM's long-context understanding capabilities, we recruited a group of experts for each of the three domains to proofread the data, and they are proficient in both English and Chinese. They need to understand the question and search for relevant evidence in multiple documents with an average length of up to 110k to judge the consistency between the question and the answer, which requires a significant amount of time and effort.

\section*{Acknowledgments}
Min Yang was supported by National Key Research and Development Program of China (2022YFF0902100), National Natural Science Foundation of China (Grant No. 62376262), the Natural Science Foundation of Guangdong Province of China (2024A1515030166), Shenzhen Science and Technology Innovation Program (KQTD20190929172835662), Shenzhen Basic Research Foundation (JCYJ20210324115614039).

\bibliography{ref}

\begin{thebibliography}{41}
\providecommand{\natexlab}[1]{#1}

\bibitem[{An et~al.(2024)An, Gong, Zhong, Zhao, Li, Zhang, Kong, and Qiu}]{an2023eval}
Chenxin An, Shansan Gong, Ming Zhong, Xingjian Zhao, Mukai Li, Jun Zhang, Lingpeng Kong, and Xipeng Qiu. 2024.
\newblock \href {https://aclanthology.org/2024.acl-long.776} {{L}-eval: Instituting standardized evaluation for long context language models}.
\newblock In \emph{Proceedings of ACL}, pages 14388--14411.

\bibitem[{Anthropic(2024{\natexlab{a}})}]{anthropic2024claude}
AI~Anthropic. 2024{\natexlab{a}}.
\newblock \href {https://www.anthropic.com/news/claude-3-family} {The claude 3 model family: Opus, sonnet, haiku}.
\newblock \emph{Claude-3 Model Card}.

\bibitem[{Anthropic(2024{\natexlab{b}})}]{anthropic2024claude35}
AI~Anthropic. 2024{\natexlab{b}}.
\newblock \href {https://www.anthropic.com/news/claude-3-5-sonnet} {Claude 3.5 sonnet model card addendum}.
\newblock \emph{Claude-3.5 Model Card}.

\bibitem[{Bai et~al.(2023)Bai, Bai, Chu, Cui, Dang, Deng, Fan, Ge, Han, Huang et~al.}]{bai2023qwen}
Jinze Bai, Shuai Bai, Yunfei Chu, Zeyu Cui, Kai Dang, Xiaodong Deng, Yang Fan, Wenbin Ge, Yu~Han, Fei Huang, et~al. 2023.
\newblock \href {https://arxiv.org/abs/2309.16609} {Qwen technical report}.
\newblock \emph{arXiv preprint arXiv:2309.16609}.

\bibitem[{Bai et~al.(2024)Bai, Lv, Zhang, Lyu, Tang, Huang, Du, Liu, Zeng, Hou, Dong, Tang, and Li}]{bai2023longbench}
Yushi Bai, Xin Lv, Jiajie Zhang, Hongchang Lyu, Jiankai Tang, Zhidian Huang, Zhengxiao Du, Xiao Liu, Aohan Zeng, Lei Hou, Yuxiao Dong, Jie Tang, and Juanzi Li. 2024.
\newblock \href {https://aclanthology.org/2024.acl-long.172} {{L}ong{B}ench: A bilingual, multitask benchmark for long context understanding}.
\newblock In \emph{Proceedings of ACL}, pages 3119--3137.

\bibitem[{bloc97(2023)}]{blocntkaware}
bloc97. 2023.
\newblock \href {https://www.reddit.com/r/LocalLLaMA/comments/14lz7j5/ntkaware_scaled_rope_allows_llama_models_to_have/} {Ntk-aware scaled rope}.
\newblock \url{https://www.reddit.com/r/LocalLLaMA/comments/14lz7j5/ntkaware_scaled_rope_allows_llama_models_to_have/}.

\bibitem[{Borgeaud et~al.(2022)Borgeaud, Mensch, Hoffmann, Cai, Rutherford, Millican, Van Den~Driessche, Lespiau, Damoc, Clark et~al.}]{borgeaud2022improving}
Sebastian Borgeaud, Arthur Mensch, Jordan Hoffmann, Trevor Cai, Eliza Rutherford, Katie Millican, George~Bm Van Den~Driessche, Jean-Baptiste Lespiau, Bogdan Damoc, Aidan Clark, et~al. 2022.
\newblock \href {https://proceedings.mlr.press/v162/borgeaud22a/borgeaud22a.pdf} {Improving language models by retrieving from trillions of tokens}.
\newblock In \emph{Proceedings of ICML}, pages 2206--2240.

\bibitem[{Chen et~al.(2024{\natexlab{a}})Chen, Lin, Han, and Sun}]{Chen_Lin_Han_Sun_2024}
Jiawei Chen, Hongyu Lin, Xianpei Han, and Le~Sun. 2024{\natexlab{a}}.
\newblock \href {https://ojs.aaai.org/index.php/AAAI/article/view/29728} {Benchmarking large language models in retrieval-augmented generation}.
\newblock In \emph{Proceedings of AAAI}, pages 17754--17762.

\bibitem[{Chen et~al.(2024{\natexlab{b}})Chen, Liu, He, Zheng, Sun, Li, Luo, and Yang}]{chen2024long}
Longze Chen, Ziqiang Liu, Wanwei He, Yinhe Zheng, Hao Sun, Yunshui Li, Run Luo, and Min Yang. 2024{\natexlab{b}}.
\newblock \href {https://aclanthology.org/2024.acl-long.447} {Long context is not long at all: A prospector of long-dependency data for large language models}.
\newblock In \emph{Proceedings of ACL}, pages 8222--8234.

\bibitem[{Chen et~al.(2023)Chen, Wong, Chen, and Tian}]{chen2023extending}
Shouyuan Chen, Sherman Wong, Liangjian Chen, and Yuandong Tian. 2023.
\newblock \href {https://arxiv.org/abs/2306.15595} {Extending context window of large language models via positional interpolation}.
\newblock \emph{arXiv preprint arXiv:2306.15595}.

\bibitem[{Dong et~al.(2024)Dong, Tang, Li, Zhao, and Wen}]{dong2023bamboo}
Zican Dong, Tianyi Tang, Junyi Li, Wayne~Xin Zhao, and Ji-Rong Wen. 2024.
\newblock \href {https://aclanthology.org/2024.lrec-main.188} {{BAMBOO}: A comprehensive benchmark for evaluating long text modeling capacities of large language models}.
\newblock In \emph{Proceedings of LREC-COLING}, pages 2086--2099.

\bibitem[{Du et~al.(2022)Du, Qian, Liu, Ding, Qiu, Yang, and Tang}]{du2022glm}
Zhengxiao Du, Yujie Qian, Xiao Liu, Ming Ding, Jiezhong Qiu, Zhilin Yang, and Jie Tang. 2022.
\newblock \href {https://aclanthology.org/2022.acl-long.26/} {Glm: General language model pretraining with autoregressive blank infilling}.
\newblock In \emph{Proceedings of ACL}, pages 320--335.

\bibitem[{Golchin and Surdeanu(2023)}]{golchin2023time}
Shahriar Golchin and Mihai Surdeanu. 2023.
\newblock \href {https://arxiv.org/abs/2308.08493} {Time travel in llms: Tracing data contamination in large language models}.
\newblock \emph{arXiv preprint arXiv:2308.08493}.

\bibitem[{Han et~al.(2024)Han, Wang, Peng, Xiong, Chen, Ji, and Wang}]{han2023lm}
Chi Han, Qifan Wang, Hao Peng, Wenhan Xiong, Yu~Chen, Heng Ji, and Sinong Wang. 2024.
\newblock \href {https://aclanthology.org/2024.naacl-long.222} {{LM}-infinite: Zero-shot extreme length generalization for large language models}.
\newblock In \emph{Proceedings of NAACL}, pages 3991--4008.

\bibitem[{Hsieh et~al.(2024)Hsieh, Sun, Kriman, Acharya, Rekesh, Jia, and Ginsburg}]{hsieh2024ruler}
Cheng-Ping Hsieh, Simeng Sun, Samuel Kriman, Shantanu Acharya, Dima Rekesh, Fei Jia, and Boris Ginsburg. 2024.
\newblock \href {https://openreview.net/forum?id=kIoBbc76Sy} {{RULER}: What{\textquoteright}s the real context size of your long-context language models?}
\newblock In \emph{Proceedings of COLM}.

\bibitem[{Kamradt(2023)}]{needleinhaystack}
Greg Kamradt. 2023.
\newblock \href {https://github.com/gkamradt/LLMTest_NeedleInAHaystack} {Needle in a haystack - pressure testing llms}.
\newblock \url{https://github.com/gkamradt/LLMTest_NeedleInAHaystack}.

\bibitem[{Karpinska et~al.(2024)Karpinska, Thai, Lo, Goyal, and Iyyer}]{karpinska2024thousandpairsnovelchallenge}
Marzena Karpinska, Katherine Thai, Kyle Lo, Tanya Goyal, and Mohit Iyyer. 2024.
\newblock \href {https://arxiv.org/abs/2406.16264} {One thousand and one pairs: A "novel" challenge for long-context language models}.
\newblock \emph{arXiv preprint arXiv: 2406.16264}.

\bibitem[{Li et~al.(2024)Li, Wang, Zheng, and Zhang}]{li2023loogle}
Jiaqi Li, Mengmeng Wang, Zilong Zheng, and Muhan Zhang. 2024.
\newblock \href {https://aclanthology.org/2024.acl-long.859} {{L}oo{GLE}: Can long-context language models understand long contexts?}
\newblock In \emph{Proceedings of ACL}, pages 16304--16333.

\bibitem[{Liu et~al.(2024)Liu, Yang, Huang, Zhang, Huang, Wei, Deng, Sun, and Zhang}]{liu-etal-2024-calibrating-llm}
Yuxuan Liu, Tianchi Yang, Shaohan Huang, Zihan Zhang, Haizhen Huang, Furu Wei, Weiwei Deng, Feng Sun, and Qi~Zhang. 2024.
\newblock \href {https://aclanthology.org/2024.lrec-main.237/} {Calibrating {LLM}-based evaluator}.
\newblock In \emph{Proceedings of LREC-COLING}, pages 2638--2656.

\bibitem[{OpenAI(2023)}]{openai2024gpt4}
OpenAI. 2023.
\newblock \href {https://arxiv.org/abs/2303.08774} {Gpt-4 technical report}.
\newblock \emph{arXiv preprint arXiv:2303.08774}.

\bibitem[{Pal et~al.(2023)Pal, Karkhanis, Roberts, Dooley, Sundararajan, and Naidu}]{pal2023giraffe}
Arka Pal, Deep Karkhanis, Manley Roberts, Samuel Dooley, Arvind Sundararajan, and Siddartha Naidu. 2023.
\newblock \href {https://arxiv.org/abs/2308.10882} {Giraffe: Adventures in expanding context lengths in llms}.
\newblock \emph{arXiv preprint arXiv:2308.10882}.

\bibitem[{Peng et~al.(2024)Peng, Quesnelle, Fan, and Shippole}]{peng2023yarn}
Bowen Peng, Jeffrey Quesnelle, Honglu Fan, and Enrico Shippole. 2024.
\newblock \href {https://openreview.net/forum?id=wHBfxhZu1u} {Ya{RN}: Efficient context window extension of large language models}.
\newblock In \emph{Proceedings of ICLR}.

\bibitem[{Press et~al.(2022)Press, Smith, and Lewis}]{press2021train}
Ofir Press, Noah Smith, and Mike Lewis. 2022.
\newblock \href {https://openreview.net/forum?id=R8sQPpGCv0} {Train short, test long: Attention with linear biases enables input length extrapolation}.
\newblock In \emph{Proceedings of ICLR}.

\bibitem[{Qiu et~al.(2024)Qiu, Li, Huang, Zhong, and King}]{qiu2024clongeval}
Zexuan Qiu, Jingjing Li, Shijue Huang, Wanjun Zhong, and Irwin King. 2024.
\newblock \href {https://arxiv.org/abs/2403.03514} {Clongeval: A chinese benchmark for evaluating long-context large language models}.
\newblock \emph{arXiv preprint arXiv:2403.03514}.

\bibitem[{Ratner et~al.(2023)Ratner, Levine, Belinkov, Ram, Magar, Abend, Karpas, Shashua, Leyton-Brown, and Shoham}]{ratner2022parallel}
Nir Ratner, Yoav Levine, Yonatan Belinkov, Ori Ram, Inbal Magar, Omri Abend, Ehud Karpas, Amnon Shashua, Kevin Leyton-Brown, and Yoav Shoham. 2023.
\newblock \href {https://aclanthology.org/2023.acl-long.352} {Parallel context windows for large language models}.
\newblock In \emph{Proceedings of ACL}, pages 6383--6402.

\bibitem[{Reid et~al.(2024)Reid, Savinov, Teplyashin, Lepikhin, Lillicrap, Alayrac, Soricut, Lazaridou, Firat, Schrittwieser et~al.}]{reid2024gemini}
Machel Reid, Nikolay Savinov, Denis Teplyashin, Dmitry Lepikhin, Timothy Lillicrap, Jean-baptiste Alayrac, Radu Soricut, Angeliki Lazaridou, Orhan Firat, Julian Schrittwieser, et~al. 2024.
\newblock \href {https://arxiv.org/abs/2403.05530} {Gemini 1.5: Unlocking multimodal understanding across millions of tokens of context}.
\newblock \emph{arXiv preprint arXiv:2403.05530}.

\bibitem[{Roziere et~al.(2023)Roziere, Gehring, Gloeckle, Sootla, Gat, Tan, Adi, Liu, Remez, Rapin et~al.}]{roziere2023code}
Baptiste Roziere, Jonas Gehring, Fabian Gloeckle, Sten Sootla, Itai Gat, Xiaoqing~Ellen Tan, Yossi Adi, Jingyu Liu, Tal Remez, J{\'e}r{\'e}my Rapin, et~al. 2023.
\newblock \href {https://arxiv.org/abs/2308.12950} {Code llama: Open foundation models for code}.
\newblock \emph{arXiv preprint arXiv:2308.12950}.

\bibitem[{Shi et~al.(2024)Shi, Min, Yasunaga, Seo, James, Lewis, Zettlemoyer, and Yih}]{shi2023replug}
Weijia Shi, Sewon Min, Michihiro Yasunaga, Minjoon Seo, Richard James, Mike Lewis, Luke Zettlemoyer, and Wen-tau Yih. 2024.
\newblock \href {https://aclanthology.org/2024.naacl-long.463} {{REPLUG}: Retrieval-augmented black-box language models}.
\newblock In \emph{Proceedings of NAACL}, pages 8371--8384.

\bibitem[{Song et~al.(2024)Song, Zheng, and Luo}]{song2024counting}
Mingyang Song, Mao Zheng, and Xuan Luo. 2024.
\newblock \href {https://arxiv.org/abs/2403.11802} {Counting-stars: A simple, efficient, and reasonable strategy for evaluating long-context large language models}.
\newblock \emph{arXiv preprint arXiv:2403.11802}.

\bibitem[{Su et~al.(2024)Su, Ahmed, Lu, Pan, Bo, and Liu}]{su2024roformer}
Jianlin Su, Murtadha Ahmed, Yu~Lu, Shengfeng Pan, Wen Bo, and Yunfeng Liu. 2024.
\newblock \href {https://dl.acm.org/doi/10.1016/j.neucom.2023.127063} {Roformer: Enhanced transformer with rotary position embedding}.
\newblock \emph{Neurocomputing}, 568:127063.

\bibitem[{Sun et~al.(2023)Sun, Dong, Patra, Ma, Huang, Benhaim, Chaudhary, Song, and Wei}]{sun2022length}
Yutao Sun, Li~Dong, Barun Patra, Shuming Ma, Shaohan Huang, Alon Benhaim, Vishrav Chaudhary, Xia Song, and Furu Wei. 2023.
\newblock \href {https://aclanthology.org/2023.acl-long.816} {A length-extrapolatable transformer}.
\newblock In \emph{Proceedings of ACL}, pages 14590--14604.

\bibitem[{Vaswani et~al.(2017)Vaswani, Shazeer, Parmar, Uszkoreit, Jones, Gomez, Kaiser, and Polosukhin}]{vaswani2017attention}
Ashish Vaswani, Noam Shazeer, Niki Parmar, Jakob Uszkoreit, Llion Jones, Aidan~N Gomez, {\L}ukasz Kaiser, and Illia Polosukhin. 2017.
\newblock \href {https://papers.nips.cc/paper_files/paper/2017/hash/3f5ee243547dee91fbd053c1c4a845aa-Abstract.html} {Attention is all you need}.
\newblock In \emph{Proceedings of NeurIPs}, page~30.

\bibitem[{Wang et~al.(2024)Wang, Cheng, Guo, Yue, Ding, Xu, Wang, Hu, Zhang, and Zhang}]{wang2024evaluating}
Cunxiang Wang, Sirui Cheng, Qipeng Guo, Yuanhao Yue, Bowen Ding, Zhikun Xu, Yidong Wang, Xiangkun Hu, Zheng Zhang, and Yue Zhang. 2024.
\newblock \href {https://proceedings.neurips.cc/paper_files/paper/2023/file/f323d594aa5d2c68154433a131c07959-Paper-Datasets_and_Benchmarks.pdf} {Evaluating open-qa evaluation}.
\newblock In \emph{Proceedings of NeurIPs}, page~36.

\bibitem[{Wu et~al.(2024)Wu, Wu, and Zou}]{wu2024faithful}
Kevin Wu, Eric Wu, and James Zou. 2024.
\newblock \href {https://arxiv.org/abs/2404.10198} {How faithful are rag models? quantifying the tug-of-war between rag and llms' internal prior}.
\newblock \emph{arXiv preprint arXiv:2404.10198}.

\bibitem[{Xiao et~al.(2023)Xiao, Tian, Chen, Han, and Lewis}]{xiao2023efficient}
Guangxuan Xiao, Yuandong Tian, Beidi Chen, Song Han, and Mike Lewis. 2023.
\newblock \href {https://arxiv.org/abs/2309.17453} {Efficient streaming language models with attention sinks}.
\newblock \emph{arXiv preprint arXiv:2309.17453}.

\bibitem[{Xiong et~al.(2024)Xiong, Liu, Molybog, Zhang, Bhargava, Hou, Martin, Rungta, Sankararaman, Oguz, Khabsa, Fang, Mehdad, Narang, Malik, Fan, Bhosale, Edunov, Lewis, Wang, and Ma}]{xiong2023effective}
Wenhan Xiong, Jingyu Liu, Igor Molybog, Hejia Zhang, Prajjwal Bhargava, Rui Hou, Louis Martin, Rashi Rungta, Karthik~Abinav Sankararaman, Barlas Oguz, Madian Khabsa, Han Fang, Yashar Mehdad, Sharan Narang, Kshitiz Malik, Angela Fan, Shruti Bhosale, Sergey Edunov, Mike Lewis, Sinong Wang, and Hao Ma. 2024.
\newblock \href {https://aclanthology.org/2024.naacl-long.260} {Effective long-context scaling of foundation models}.
\newblock In \emph{Proceedings of NAACL}, pages 4643--4663.

\bibitem[{Xu et~al.(2024)Xu, Ping, Wu, McAfee, Zhu, Liu, Subramanian, Bakhturina, Shoeybi, and Catanzaro}]{xu2023retrieval}
Peng Xu, Wei Ping, Xianchao Wu, Lawrence McAfee, Chen Zhu, Zihan Liu, Sandeep Subramanian, Evelina Bakhturina, Mohammad Shoeybi, and Bryan Catanzaro. 2024.
\newblock \href {https://openreview.net/forum?id=xw5nxFWMlo} {Retrieval meets long context large language models}.
\newblock In \emph{Proceedings of ICLR}.

\bibitem[{Zhang et~al.(2024{\natexlab{a}})Zhang, Li, Liu, Yang, Liu, Chen, Luo, and Yang}]{zhang2023marathon}
Lei Zhang, Yunshui Li, Ziqiang Liu, Jiaxi Yang, Junhao Liu, Longze Chen, Run Luo, and Min Yang. 2024{\natexlab{a}}.
\newblock \href {https://aclanthology.org/2024.acl-long.284} {Marathon: A race through the realm of long context with large language models}.
\newblock In \emph{Proceedings of ACL}, pages 5201--5217.

\bibitem[{Zhang et~al.(2023)Zhang, Yu, Yu, Lv, Liu, Huang, Xu, and Li}]{zhang2023wider}
Xinghua Zhang, Bowen Yu, Haiyang Yu, Yangyu Lv, Tingwen Liu, Fei Huang, Hongbo Xu, and Yongbin Li. 2023.
\newblock \href {https://arxiv.org/abs/2308.01862} {Wider and deeper llm networks are fairer llm evaluators}.
\newblock \emph{arXiv preprint arXiv:2308.01862}.

\bibitem[{Zhang et~al.(2024{\natexlab{b}})Zhang, Chen, Hu, Xu, Chen, Hao, Han, Thai, Wang, Liu, and Sun}]{zhang2024infty}
Xinrong Zhang, Yingfa Chen, Shengding Hu, Zihang Xu, Junhao Chen, Moo Hao, Xu~Han, Zhen Thai, Shuo Wang, Zhiyuan Liu, and Maosong Sun. 2024{\natexlab{b}}.
\newblock \href {https://aclanthology.org/2024.acl-long.814} {$\infty${B}ench: Extending long context evaluation beyond 100{K} tokens}.
\newblock In \emph{Proceedings of ACL}, pages 15262--15277.

\bibitem[{Zhu et~al.(2024)Zhu, Yang, Wang, Song, Wu, Wei, and Li}]{zhu2023pose}
Dawei Zhu, Nan Yang, Liang Wang, Yifan Song, Wenhao Wu, Furu Wei, and Sujian Li. 2024.
\newblock \href {https://openreview.net/forum?id=3Z1gxuAQrA} {Po{SE}: Efficient context window extension of {LLM}s via positional skip-wise training}.
\newblock In \emph{Proceedings of ICLR}.

\end{thebibliography}
\clearpage
\appendix

\section{GPT4-as-the-Judge Prompt}
\label{paper:GPT4-as-the-Judge}
In Loong, GPT4 is used as a Judger to evaluate the correctness of the model-generated content, with the prompt used shown in the following. With this evaluation method, we expect the Judger model to output a percentage score along with its corresponding explanation.
\begin{tcolorbox}[colback=white]
\small
\textbf{[Gold Answer]}\\ 
\textcolor{blue}{<answer>}\\

\textbf{[The Start of Assistant's Predicted Answer]}\\
 \textcolor{blue}{<LLM's response>}\\
\textbf{[The End of Assistant's Predicted Answer]}\\

\textbf{[System]}\\
We would like to request your feedback on the performance of the AI assistant in response to the user question displayed above according to the gold answer. Please use the following listed aspects and their descriptions as evaluation criteria:\\

    - Accuracy and Hallucinations: The assistant's answer is semantically consistent with the gold answer; The numerical value and order need to be accurate, and there should be no hallucinations.\\
    - Completeness: Referring to the reference answers, the assistant's answer should contain all the key points needed to answer the user's question; further elaboration on these key points can be omitted.\\
Please rate whether this answer is suitable for the question. Please note that the gold answer can be considered as a correct answer to the question.\\

The assistant receives an overall score on a scale of 1 to 100, where a higher score indicates better overall performance.Please note that if the assistant's answer and the gold answer fully meet the above criteria, its overall rating should be the full marks (100). Please first provide a comprehensive explanation of your evaluation, avoiding any potential bias.Then, output a line indicating the score of the Assistant.\\

PLEASE OUTPUT WITH THE FOLLOWING FORMAT, WHERE THE SCORE IS A SCALE OF 1 TO 100 BY STRICTLY FOLLOWING THIS FORMAT: "[[score]]", FOR EXAMPLE "Rating: [[100]]":\\

\textbf{<Start Output>}\\
Evaluation evidence: your evluation explanation here, no more than 100 words
Rating: [[score]]\\
\textbf{<End Output>}\\

Now, start your evaluation:
\label{text:judger}
\end{tcolorbox}

\section{Test Case}
\label{paper:Case}
To facilitate understanding of Loong's data examples, we present examples of 11 sub-tasks in the following, showing the format we input to the model as well as the prompts we used.

\subsection{Spotlight Locating}
\begin{framed}
\noindent\textbf{<Multi\_Documents>}\\
\textbf{Prompt:} Please answer the following questions based only on the judgment documents you have seen above. You only need to give the titles of the judgment documents that meet the requirements.\\ 
\textbf{Question:} Among the above judgment documents, which one is the case of 'crime of endangering public security'?\\
\textbf{Answer:} Judgment Document 5
\end{framed}

\subsection{Sequential Enumeration}
\begin{framed}
\noindent\textbf{<Multi\_Documents>}\\
\textbf{Prompt:} We kindly ask you to review the financial statements of the companies provided above and answer the following questions based solely on the information you have seen. If the question involves content not found in the financial statements, you may ignore this part and only answer the other parts.\\ 
\textbf{Question:} Please list the Cash and Cash Equivalents of each of the aforementioned companies in ascending order?\\
\textbf{Answer:} \$ 1,273 thousand, \$ 1,360 thousand, \$ 9,364 thousand
\end{framed}

\subsection{Extremum Acquisition}
\begin{framed}
\noindent\textbf{<Multi\_Documents>}\\
\textbf{Prompt:} We kindly ask you to review the financial statements of the companies provided above and answer the following questions based solely on the information you have seen. If the question involves content not found in the financial statements, you may ignore this part and only answer the other parts.\\ 
\textbf{Question:} Which company has the highest 'Total Non-current Assets'?\\
\textbf{Answer:} BLUE DOLPHIN ENERGY CO with \$56,787,000
\end{framed}

\subsection{Range Awareness}
\begin{framed}
\noindent\textbf{<Multi\_Documents>}\\
\textbf{Prompt:} We kindly ask you to review the financial statements of the companies provided above and answer the following questions based solely on the information you have seen. If the question involves content not found in the financial statements, you may ignore this part and only answer the other parts.\\ 
\textbf{Question:} How many companies have 'Total Shares Outstanding' exceeding 10,000,000 shares?\\
\textbf{Answer:} 4 companies
\end{framed}

\subsection{Report Integration}
\begin{framed}
\noindent\textbf{<Multi\_Documents>}\\
\textbf{Prompt:} We kindly ask you to review the financial statements of the companies provided above and answer the following questions based solely on the information you have seen. If the question involves content not found in the financial statements, you may ignore this part and only answer the other parts.\\ 
\textbf{Question:} Please categorize the companies listed above by 'Total Shares Outstanding' into the following groups: below 10,000,000 shares and 10,000,000 shares or more. Place companies into the same collection for the same category and into different collections for different categories.\\
\textbf{Answer:} \{"below 10,000,000 shares": ["GSE SYSTEMS INC", "CROSS TIMBERS ROYALTY TRUST"], "10,000,000 shares or more": ["HUGOTON ROYALTY TRUST"]\}
\end{framed}

\subsection{Citation\&Reference}
\begin{framed}
\noindent\textbf{<Multi\_Documents>}\\
\textbf{Prompt:} We hope you will carefully study the provided papers and determine the citation relationships between them. Please follow the instructions below strictly to complete the task:\\
\noindent\#Specific Requirements:\\
1. Reference: When a given paper mentions other provided papers, those other papers are considered as "references" for the given paper. To summarize in this specific context, references are about what the given paper is using.\\
2. Citation: Conversely, when other provided papers mention the given paper in their works, the given paper is being "cited" by those other papers. To summarize in this specific context, citations are about who is using the given paper.\\
3. Given a paper, you need to determine the citation or reference relationship between this paper and the other papers. Do not consider papers that are not provided.\\
3. Please present the paper titles in a json format as follows: \{"Reference":["Reference Title 1", "Reference Title 2", ..., "Reference Title n"], "Citation":["Citation Title 1", "Citation Title 2", ..., "Citation Title n"]\}.\\
4. If a paper does not have any references or citations, please leave the corresponding list empty, e.g.\{"Refernce":[]\}, \{"Citation":[]\}.\\ 
\textbf{Question:} The paper you need to analyze:Self-Discover: Large Language Models Self-Compose Reasoning Structures\\
\textbf{Answer:} \{'Reference': ['\# Plan, Verify and Switch: Integrated Reasoning with Diverse X-of-Thoughts ', '\# StrategyLLM: Large Language Models as Strategy Generators, Executors, Optimizers, and Evaluators for Problem Solving '], 'Citation': ['\# Can LLMs Solve Longer Math Word Problems Better? ']\}
\end{framed}

\subsection{Case Classification}
\begin{framed}
\noindent\textbf{<Multi\_Documents>}\\
\textbf{Prompt:} Please answer the following questions based only on the judgment documents you have seen above. You only need to give the titles of the judgment documents that meet the requirements.\\
\textbf{Question:} After reading the above judgments, please classify all the judgments according to the following three types of cases: 'Civil Cases', 'Enforcement Cases', and 'Administrative Cases'.\\
\textbf{Answer:}  \{"Civil Cases": ["Judgment Document 2"], "Enforcement Cases": ["Judgment Document 4"], "Administrative Cases": ["Judgment Document 1", "Judgment Document 3"]\}
\end{framed}

\subsection{Temporal Analysis}
\begin{framed}
\noindent\textbf{<Multi\_Documents>}\\
\textbf{Prompt:} We kindly ask you to review the financial statements of the companies provided above and answer the following questions based solely on the information you have seen. If the question involves content not found in the financial statements, you may ignore this part and only answer the other parts.\\
\textbf{Question:} What is the trend in ARVANA INC's share capital from 2021 to 2024?\\
\textbf{Answer:} ARVANA INC's share capital has consistently increased from \$4,611 in 2021 to \$34,149 in 2022, \$35,949 in 2023, and \$107,847 in 2024.
\end{framed}

\subsection{Citation Chain}
\begin{framed}
\noindent\textbf{<Multi\_Documents>}\\
\textbf{Prompt:} We kindly ask you to thoroughly review the provided papers and construct a citation chain from them. Please adhere to the following instructions strictly while completing the task:\\
\noindent\#Task Instructions:\\
\noindent Given several papers, you are required to identify and list the longest citation chain, which demonstrates the citation relationship among the provided papers.\\
\noindent\#Specific Requirements:\\
\noindent1.Please present the titles of the papers in the form of a list, as follows: ["Title of Paper 1", "Title of Paper 2", ..., "Title of Paper n"].\\
2.Ensure that the citation chain in the list is linear and continuous, meaning that the first paper title in the list (Paper 1) should not cite any other works. Instead, it should be cited by the next paper in the list (Paper 2); subsequently, each paper should then be cited by the next one in the list, continuing up to the last paper (Paper n).\\
3.Consider only the citation relationships within the supplied collection of papers, and ensure that the citation chain accurately reflects the sequential citation order among these documents.\\
4.Do not take into account any articles not provided, and disregard other non-linear citation relationships.
\\
\textbf{Answer:} ["\# Very Deep Transformers for Neural Machine Translation ", "\# Understanding the Difficulty of Training Transformers ", "\# MonaCoBERT: Monotonic attention based ConvBERT for Knowledge Tracing"]
\end{framed}

\subsection{Link the Links}
\begin{framed}
\noindent\textbf{<Multi\_Documents>}\\
\textbf{Prompt:} Answer the following questions based solely on the judgment document you have seen above.\\
\textbf{Question:}After reading the above judgment document, I will give you several judgment results: \textcolor{black}{<a list of judgment result: judgment1\textasciitilde judgment6>} You need to determine the most likely judgment result for each of the above judgment documents.\\
\textbf{Answer:} \{"Judgment Document 1": "Judgment Result 1", "Judgment Document 2": "Judgment Result 6", "Judgment Document 3": "Judgment Result 2", "Judgment Document 4": "Judgment Result 5"\}
\end{framed}

\subsection{Solitaire}
\begin{framed}
\noindent\textbf{<Multi\_Documents>}\\
\textbf{Prompt:} Answer the following questions based solely on the judgment document you have seen above.\\
\textbf{Question:}
Reading the above judgments, I will provide several case types arranged in a left-to-right sequence: ['CaseType1', 'CaseType2', 'CaseType3', 'CaseType4']. You need to sort all the judgment documents according to the above sequence of case types. The judgment documents only need to include the titles.\\
Please provide the answer:\\
\textbf{Answer:} \{"CaseType1": "Judgment Document 3", "CaseType2": "Judgment Document 1", "CaseType3": "Judgment Document 4", "CaseType4": "Judgment Document 6"\}
\end{framed}

\section{Length Distribution}
\label{paper:length}
\begin{figure}[t]
    \centering
    \includegraphics[width=1.0\columnwidth]{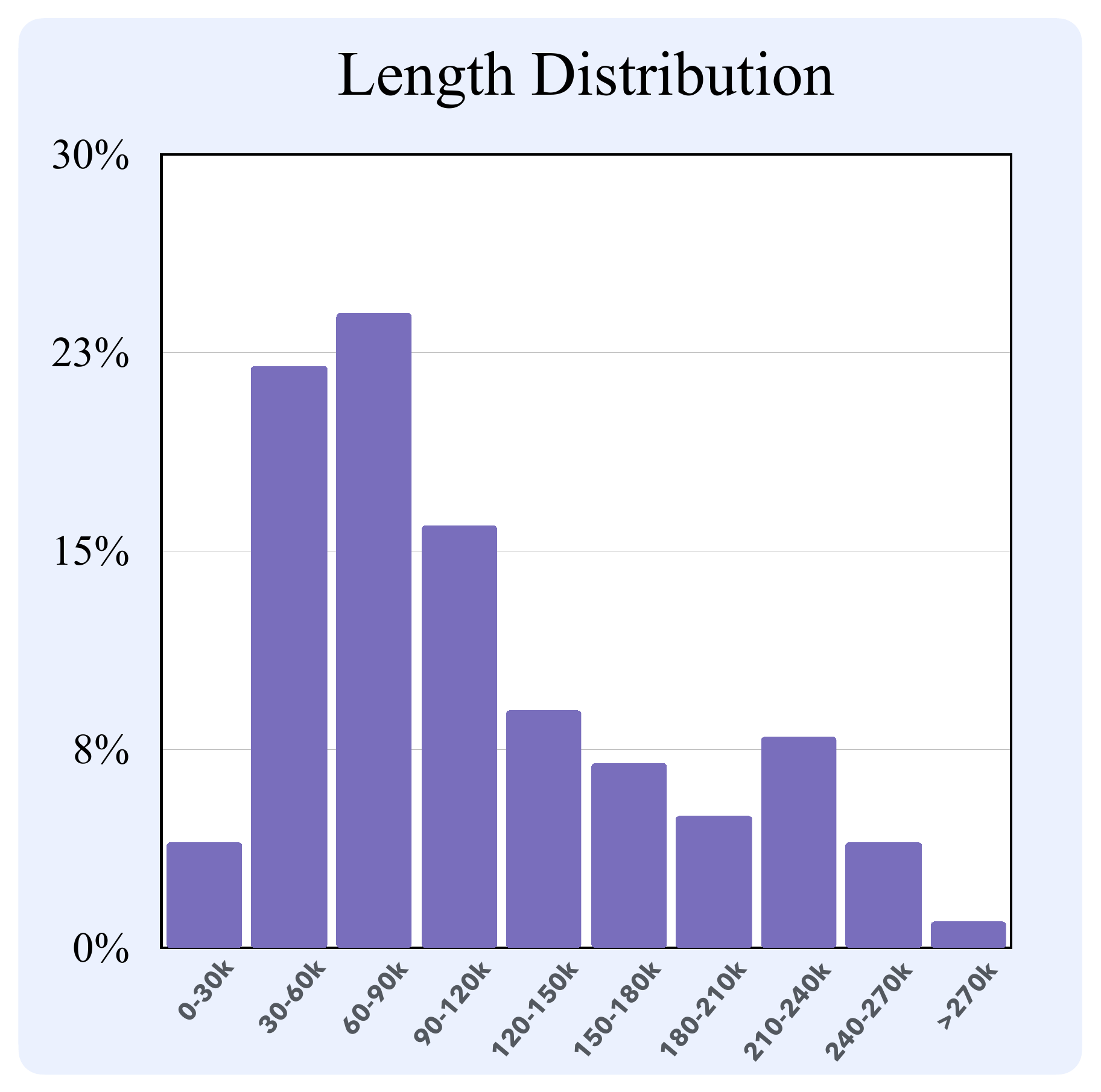}
    \caption{Test Case Length Distribution in Loong.}
    \label{fig:length}
\end{figure}

As shown in \Cref{fig:length} and \Cref{tb:app_stat}, we present the distribution of data lengths in Loong. It can be observed that the data is primarily distributed around 30-150k. Moreover, we have sufficient data in both shorter and longer ranges, allowing us to assess the model's capabilities across each length interval.

\section{RAG Detailed Results}
\label{paper:rag}
We conducted experiments on GPT-4o and Qwen2-72B-Instruct with the addition of a RAG module. As shown in \Cref{tb:rag_overall}, \Cref{tb:rag_gpt4}, and \Cref{tb:rag_qwen2}, we have published detailed experimental results. It can be seen that RAG achieved subpar results on our Loong, indicating that Loong requires the model to have genuine long-context understanding capabilities.

\section{Comparison of Evidence Distribution}
\label{paper:evidence_distribution}
In the same multi-document question-answering task, we compared the distribution of evidence related to the answers in the context for Loong (Ours) and Longbench~\cite{bai2023longbench}. As shown in \Cref{table:evidence}, we present data examples from Loong and LongBench. It can be observed that although Longbench contains a large number of passages, the evidence is only distributed within Passage 1. In contrast, in our Loong, the evidence is distributed across every document, requiring the model to understand each document in order to provide the correct answer.

\section{Comparison of Results with Other Benchmarks}
\label{paper: comparison_benchmark}
Due to the rapid iteration of models and the high cost of closed-source model APIs, previous studies have rarely evaluated models that support context windows of 128k or more.  Evaluating the latest models with ultra-long context windows and expensive closed-source models is our advantage compared to previous methods. Users can evaluate the gap between their models and the latest open-source and closed-source models at a low cost. To discover more interesting conclusions, we compared Loong with RULER ~\cite{hsieh2024ruler}, and the most recent work, NOCHA ~\cite{karpinska2024thousandpairsnovelchallenge}. The experimental results are shown in the \Cref{tb:r_b}. 

From the result, we discovered that: (1) Although RULER focuses on synthetic tasks and NOCHA focuses on the domain of novels, the performance of the models on the three benchmarks is essentially consistent. (2) Gemini-1.5-pro achieved the best results in Loong and RULER, but GPT-4o had a significant lead on NOCHA. We believe that since NOCHA's data sources only include novels, GPT-4o might have stronger capabilities in the domain of novel literature. (3) The large-parameter Qwen2-72B-Instruct underperformed in RULER compared to the smaller-parameter models GLM4-9B-Chat and Llama3.1-8B, which is inconsistent with the results of Loong. We believe that RULER emphasizes synthetic tasks, where the performance of such tasks is as sensitive as Needle In A Haystack. Specifically, simply adding some synthetic task data to the training corpus can yield good results. In real-world tasks like Loong, such anomalous phenomena generally do not occur.
\begin{table}[t]
\centering  
\resizebox{1\columnwidth}{!}{
\begin{tabular}{lccc}
\toprule
Models & Loong & RULER & NOCHA\\
\midrule
GPT-4o & 53.47 & - & 55.80\\
GPT-4-Turbo & -	& 91.60 & 40.20\\
Gemini-1.5-pro & 55.37 & 95.80 & 48.10\\
Claude3.5-Sonnet & 48.85 & - & 41.00\\
Qwen2-72B-Instruct & 43.29 & 85.90 & 43.35\\
GLM4-9B-Chat & 38.31 & 89.90 & 27.07\\
Llama3.1-8B	& 36.31 & 88.30 & 16.53\\
Phi-3-mini-3.8B & 14.54 & 68.80 & 9.30\\
\bottomrule
\end{tabular}
}
\caption{Compare Loong with RULER and NOCHA.}
\label{tb:r_b}
\end{table}

\section{Results of Recall Rate by RAG}
\label{paper:RAG_recall_rate}
To show that RAG does not cover all evidence, we sampled all the questions from the financial test case in the Loong and analyzed RAG‘s retrieval results. Since the answers to our questions are distributed across all documents on average, we evaluated whether the retrieved top-k passages cover all documents to reflect the RAG's recall of evidence. \textit{This metric will be higher than the actual evidence recall rate because even if all required documents are retrieved, the specific passages may not necessarily contain the evidence. If the model fails to retrieve all documents, it will certainly be unable to retrieve all evidence.} The evaluation metric used is \textit{Recall@n}, where n represents the number of retrieved passages. If the n passages contain all documents, \textit{Recall@n} is set to 1; otherwise, it is 0. The results can be seen in \Cref{tb:recall_rate}. Evidently, in the Loong dataset, RAG struggles to retrieve all relevant documents. Even when the topK reaches 50, the highest recall is only 0.64. In reality, the recall rate for all evidence will be even lower.

\begin{table*}[t]
\centering  
\resizebox{\textwidth}{!}{
\begin{tabular}{lccccc}
\toprule
Model & Recall@10 & Recall@20 & Recall@30 & Recall@40 & Recall@50\\
\midrule
Openai Embedding, chunk size=1024 & 0.20 & 0.34 & 0.43 & 0.51 & 0.58\\
Openai Embedding, chunk size=2048 & 0.20 & 0.36 & 0.48 & 0.57 & 0.62\\
BGE Embedding, chunk size=1024 & 0.19 & 0.34 & 0.44 & 0.53 & 0.59\\
BGE Embedding, chunk size=2048 & 0.21 & 0.38 & 0.49 & 0.58 & 0.64\\
\bottomrule
\end{tabular}
}
\caption{The Recall Rate of all documents by using RAG.}
\label{tb:recall_rate}
\end{table*}

\section{Detailed URL of Document Source}
\label{paper:doc_url}
China Judge Online: 

\url{https://wenshu.court.gov.cn/};

\noindent U.S. SEC: 

\url{https://www.sec.gov/};

\noindent cninf: 

\url{http://www.cninfo.com.cn/};

\noindent Arxiv: 

\url{https://arxiv.org/};

\noindent Semantic Scholar: 

\url{https://www.semanticscholar.org/}

\begin{table*}[htbp]
\centering  
\resizebox{\textwidth}{!}{
\begin{tabular}{lcccc}
\toprule
Dataset & \#data in 10-50k & \#data in 50-100K & \#data in 100K-200K & \#data in 200-250K\\
\midrule
\emph{Spotlight Locating} & \emph{53} & \emph{70} & \emph{80} & \emph{47}\\
\midrule
\emph{Comparison} & \emph{60} & \emph{105} & \emph{95} & \emph{40}\\
Sequential Enumeration & 24 & 29 & 20 & 14\\
Extremum Acquisition & 16 & 55 & 59 & 13\\
Range Awareness & 20 & 21 & 16 & 13\\
\midrule
\emph{Clustering} & \emph{113} & \emph{246} & \emph{194} & \emph{88}\\
Report Integration & 40 & 90 & 90 & 30\\
Citation\&Reference & 37 & 120 & 79 & 34\\
Case Classification & 36 & 36 & 25 & 24\\
\midrule
\emph{Chain of Reasoning} & \emph{97} & \emph{143} & \emph{112} & \emph{57}\\
Temporal Analysis & 10 & 40 & 35 & 15\\
Citation Chain & 33 & 50 & 41 & 6\\
Link the Links & 35 & 25 & 28 & 25\\
Solitaire & 28 & 19 & 8 & 11\\
\midrule
\emph{Overall} & \emph{323} & \emph{564} & \emph{281} & \emph{232}\\
Chinese & 240 & 284 & 251 & 130\\
English & 83 & 280 & 230 & 102\\
\bottomrule
\end{tabular}
}
\caption{Data length distributions in Loong benchmark.}
\label{tb:app_stat}
\end{table*}
\begin{table*}[t]
\centering  
\resizebox{\textwidth}{!}{
\begin{tabular}{l|cc|cc|cc|cc|cc}
\toprule
\multirow{1}{*}{\textbf{Model}} & \multicolumn{2}{c|}{\textbf{Spotlight Locating}} & \multicolumn{2}{c|}{\textbf{Comparison}} & \multicolumn{2}{c|}{\textbf{Clustering}} & \multicolumn{2}{c|}{\textbf{Chain of Reasoning}} & \multicolumn{2}{c}{\textbf{Overall}}\\
\midrule
GPT4o (128K) & 73.95 &0.62 & 50.50 &0.28 &44.29 &0.09 &57.95 &0.28 &53.47 &0.26 \\
\textit{w/ Openai Embedding, Top k=5} & 56.97 &0.36 &31.28 &0.14 &27.71 &0.03 &26.65 &0.04 &32.85 &0.11 \\
\textit{w/ BGE Embedding, Top k=5} &63.32 &0.44 &34.63 &0.17 &26.74 &0.03 &26.21 &0.04 &34.01 &0.13 \\
\textit{w/ Openai Embedding, Top k=10} & 65.20 &0.46 &36.80 &0.19 &33.06 &0.04 &33.26 &0.08 &38.80 &0.14 \\
\textit{w/ BGE Embedding, Top k=10}& 68.27 &0.50 &39.51 &0.22 &31.91 &0.04 &30.71 &0.07 &38.71 &0.15 \\
\textit{w/ Openai Embedding, Top k=30} &64.32 &0.43 &42.15 &0.26 &41.02 &0.08 &40.14 &0.14 &44.62 &0.18 \\
\textit{w/ BGE Embedding, Top k=30}&64.76 &0.45 &47.56 &0.32 &40.43 &0.08 &40.82 &0.17 &45.67 &0.21 \\
\textit{w/ Openai Embedding, Top k=50} & 65.59 &0.45 &41.69 &0.28 &34.49 &0.04 &39.74 &0.14 &42.70 &0.18 \\
\textit{w/ BGE Embedding, Top k=50}&63.28  &0.42 &47.05 &0.32 &42.64 &0.10 &41.97 &0.18 &46.52 &0.21 \\
\midrule
Qwen2-72B-Instruct (128K) & 54.17 & 0.36 &42.38 & 0.20 & 36.71 & 0.04 & 47.76 & 0.18 & 43.29 & 0.15\\ 
\textit{w/ Openai Embedding, Top k=5} &57.57  &0.40 &30.98 &0.14 &27.91 &0.02 &28.30 &0.04 &33.22 &0.10 \\
\textit{w/ BGE Embedding, Top k=5} & 62.02 &0.44 &32.90 &0.16 &27.05 &0.02 &29.26 &0.06 &34.18 &0.12 \\
\textit{w/ Openai Embedding, Top k=10} & 62.52 &0.44 &35.79 &0.18 &30.16 &0.03 &32.67 &0.08 &36.92 &0.13 \\
\textit{w/ BGE Embedding, Top k=10}& 69.24 &0.51 &36.78 &0.18 &29.07 &0.02 &31.90 &0.07 &37.50 &0.14 \\
\textit{w/ Openai Embedding, Top k=30}&64.11 &0.43 &41.91 &0.26 &35.61 &0.04 &42.61 &0.19 &42.98 &0.18 \\
\textit{w/ BGE Embedding, Top k=30}&58.59 &0.38 &42.66 &0.23 &33.77 &0.05 &40.39 &0.17 &40.94 &0.16 \\
\textit{w/ Openai Embedding, Top k=50} &57.87 &0.37 &42.10 &0.25 &32.78 &0.03 &42.39 &0.19 &40.86 &0.17 \\
\textit{w/ BGE Embedding, Top k=50} &56.93 &0.37 &39.51 &0.20 &32.67 &0.04 &43.44 &0.21 &40.46 &0.16 \\

\bottomrule
\end{tabular}
}
\caption{Overall results (\%) of adding RAG module on GPT4o and Qwen2-72B-Instruct.}
\label{tb:rag_overall}
\end{table*}
\begin{table*}[t]
\centering  
\resizebox{\textwidth}{!}{
\begin{tabular}{l|cc|cc|cc|cc|cc}
\toprule
\multirow{1}{*}{\textbf{Model}} & \multicolumn{2}{c|}{\textbf{Spotlight Locating}} & \multicolumn{2}{c|}{\textbf{Comparison}} & \multicolumn{2}{c|}{\textbf{Clustering}} & \multicolumn{2}{c|}{\textbf{Chain of Reasoning}} & \multicolumn{2}{c}{\textbf{Overall}}\\
\midrule
\multicolumn{11}{c}{\textbf{ $\mathtt{Set1}$ (10K-50K)}} \\
GPT-4o (128K) & 85.67 &0.81 & 64.27 &0.33 &57.01 &0.24 &81.58 &0.55 &70.40 &0.44 \\
\textit{w/ Openai Embedding, Top k=5} & 47.60 &0.31 &29.75 &0.10 &29.10 &0.06 &31.46 &0.08 &32.98 &0.11 \\
\textit{w/ BGE Embedding, Top k=5} &57.17 &0.43 &34.15 &0.12 &30.71 &0.07 &28.77 &0.08 &35.23 &0.14 \\
\textit{w/ Openai Embedding, Top k=10} &61.25 &0.44 &38.33 &0.17 &37.00 &0.08 &41.67 &0.16 &42.63 &0.18 \\
\textit{w/ BGE Embedding, Top k=10}&61.00 &0.44 &39.74 &0.19 &36.14 &0.08 &34.90 &0.11 &40.44 &0.17 \\
\textit{w/ Openai Embedding, Top k=30} &55.15 &0.37 &46.60 &0.28 &45.54 &0.13 &51.98 &0.27 &49.23 &0.24 \\
\textit{w/ BGE Embedding, Top k=30}&57.40 &0.38 &52.25 &0.32 &46.54 &0.18 &50.02 &0.25 &50.41 &0.26 \\
\textit{w/ Openai Embedding, Top k=50} &55.47 &0.40 &49.62 &0.33 &39.61 &0.10 &46.08 &0.20 &46.82 &0.24 \\
\textit{w/ BGE Embedding, Top k=50}&52.08 &0.38 &53.42 &0.37 &49.83 &0.21 &48.88 &0.24 &50.55 &0.27 \\

\midrule
\multicolumn{11}{c}{\textbf{ $\mathtt{Set2}$ (50K-100K)}} \\
GPT-4o (128K) & 86.76 &0.72 & 59.81 &0.40 &47.83 &0.11 &62.09 &0.34 &58.38 &0.29 \\
\textit{w/ Openai Embedding, Top k=5} & 56.01 &0.35 &39.56 &0.22 &31.84 &0.04 &27.01 &0.03 &35.31 &0.11 \\
\textit{w/ BGE Embedding, Top k=5} &67.33 &0.43 &43.90 &0.28 &29.37 &0.04 &27.84 &0.04 &36.72 &0.14 \\
\textit{w/ Openai Embedding, Top k=10} &64.77 &0.45 &45.44 &0.31 &36.07 &0.05 &32.29 &0.05 &40.54 &0.15 \\
\textit{w/ BGE Embedding, Top k=10}&72.07 &0.52 &50.15 &0.32 &34.35 &0.05 &33.49 &0.07 &41.90 &0.17 \\
\textit{w/ Openai Embedding, Top k=30} &65.87 &0.42 &50.05 &0.34 &44.08 &0.11 &42.60 &0.15 &47.48 &0.20 \\
\textit{w/ BGE Embedding, Top k=30}&65.26 &0.49 &55.21 &0.43 &43.82 &0.10 &42.80 &0.18 &48.31 &0.23 \\
\textit{w/ Openai Embedding, Top k=50} &67.21 &0.46 &51.38 &0.38 &31.65 &0.03 &40.69 &0.12 &43.52 &0.19 \\
\textit{w/ BGE Embedding, Top k=50} &67.43 &0.46 &53.98 &0.39 &45.04 &0.12 &46.94 &0.21 &49.96 &0.24 \\

\midrule
\multicolumn{11}{c}{\textbf{ $\mathtt{Set3}$ (100K-200K)}} \\
GPT-4o (128K) & 74.84 &0.65 & 42.40 &0.21 &38.70 &0.04 &45.06 &0.09 &46.95 &0.19 \\
\textit{w/ Openai Embedding, Top k=5} & 67.45 &0.49 &29.00 &0.13 &25.09 &0.01 &27.22 &0.02 &33.69 &0.12 \\
\textit{w/ BGE Embedding, Top k=5} &71.12 &0.56 &31.36 &0.14 &25.32 &0.00 &25.78 &0.04 &34.43 &0.13 \\
\textit{w/ Openai Embedding, Top k=10} &72.37 &0.55 &31.41 &0.13 &30.59 &0.01 &33.14 &0.08 &38.38 &0.14 \\
\textit{w/ BGE Embedding, Top k=10}&79.04 &0.67 &34.29 &0.18 &30.59 &0.02 &29.69 &0.06 &39.22 &0.17 \\
\textit{w/ Openai Embedding, Top k=30} &74.03 &0.57 &37.00 &0.22 &39.53 &0.04 &36.07 &0.09 &43.91 &0.18 \\
\textit{w/ BGE Embedding, Top k=30}&75.45 &0.59 &45.96 &0.31 &36.91 &0.04 &39.16 &0.14 &45.68 &0.21 \\
\textit{w/ Openai Embedding, Top k=50} &77.24 &0.59 &25.02 &0.12 &37.55 &0.05 &40.93 &0.15 &44.52 &0.19 \\
\textit{w/ BGE Embedding, Top k=50} &74.19 &0.55 &46.15 &0.31 &39.71 &0.04 &36.56 &0.15 &45.99 &0.20 \\

\midrule
\multicolumn{11}{c}{\textbf{ $\mathtt{Set4}$ (200K-250K)}} \\
GPT-4o (128K) & 36.79 &0.19 & 23.97 &0.08 &30.40 &0.00 &32.89 &0.07 &31.11 &0.07 \\
\textit{w/ Openai Embedding, Top k=5} & 50.76 &0.22 &17.25 &0.00 &19.53 &0.00 &16.61 &0.00 &24.91 &0.05 \\
\textit{w/ BGE Embedding, Top k=5} &51.02 &0.26 &18.75 &0.03 &17.83 &0.00 &18.77 &0.02 &25.07 &0.06 \\
\textit{w/ Openai Embedding, Top k=10} &57.98 &0.31 &23.00 &0.03 &25.08 &0.00 &21.29 &0.02 &30.00 &0.07 \\
\textit{w/ BGE Embedding, Top k=10}&51.48 &0.25 &23.36 &0.05 &22.55 &0.00 &18.95 &0.02 &27.48 &0.06 \\
\textit{w/ Openai Embedding, Top k=30} &55.85 &0.26 &26.38 &0.08 &29.94 &0.00 &21.81 &0.02 &32.66 &0.07 \\
\textit{w/ BGE Embedding, Top k=30}&53.94 &0.21 &24.82 &0.07 &30.77 &0.00 &23.61 &0.05 &32.68 &0.07 \\
\textit{w/ Openai Embedding, Top k=50} &55.00 &0.28 &12.79 &0.04 &29.64 &0.00 &24.67 &0.07 &31.88 &0.08 \\
\textit{w/ BGE Embedding, Top k=50} &51.17 &0.21 &23.36 &0.10 &33.08 &0.02 &28.39 &0.09 &33.82 &0.09 \\

\bottomrule
\end{tabular}
}
\caption{The result of adding RAG module on GPT-4o with different length sets.}
\label{tb:rag_gpt4}
\end{table*}
\begin{table*}[t]
\centering  
\resizebox{\textwidth}{!}{
\begin{tabular}{l|cc|cc|cc|cc|cc}
\toprule

\multirow{1}{*}{\textbf{Model}} & \multicolumn{2}{c|}{\textbf{Spotlight Locating}} & \multicolumn{2}{c|}{\textbf{Comparison}} & \multicolumn{2}{c|}{\textbf{Clustering}} & \multicolumn{2}{c|}{\textbf{Chain of Reasoning}} & \multicolumn{2}{c}{\textbf{Overall}}\\
\midrule
\multicolumn{11}{c}{\textbf{ $\mathtt{Set1}$ (10K-50K)}} \\
Qwen2-72B-Instruct (128K) & 68.49 & 0.55 & 60.60 & 0.37 & 47.08 & 0.08 & 70.39 & 0.36 & 60.11 & 0.29\\ 
\textit{w/ Openai Embedding, Top k=5} &54.62 &0.45 &26.17 &0.08 &29.60 &0.03 &34.41 &0.08 &34.51 &0.12 \\
\textit{w/ BGE Embedding, Top k=5} & 62.92 &0.53 &30.92 &0.08 &31.28 &0.03 &32.95 &0.11 &36.91 &0.15 \\
\textit{w/ Openai Embedding, Top k=10} & 59.81 &0.43 &34.93 &0.15 &29.33 &0.02 &41.27 &0.15 &38.96 &0.15 \\
\textit{w/ BGE Embedding, Top k=10}& 72.13 &0.62 &32.42 &0.12 &31.90 &0.05 &44.12 &0.20 &42.27 &0.20 \\
\textit{w/ Openai Embedding, Top k=30}&57.26 &0.40 &45.43 &0.28 &40.04 &0.06 &57.32 &0.35 &49.06 &0.24 \\
\textit{w/ BGE Embedding, Top k=30}&56.37 &0.33 &46.27 &0.30 &38.35 &0.10 &51.49 &0.29 &46.69 &0.23 \\
\textit{w/ Openai Embedding, Top k=50} &51.08 &0.35 &44.53 &0.27 &37.96 &0.05 &53.95 &0.35 &46.11 &0.23 \\
\textit{w/ BGE Embedding, Top k=50}&53.47 &0.37 &47.31 &0.29 &36.42 &0.06 &54.65 &0.35 &46.62 &0.24 \\

\midrule
\multicolumn{11}{c}{\textbf{ $\mathtt{Set2}$ (50K-100K)}} \\
Qwen2-72B-Instruct (128K) & 64.53 & 0.43 & 42.60 & 0.21 & 38.52 & 0.05 & 51.18 & 0.20 & 45.71 & 0.17\\ 
\textit{w/ Openai Embedding, Top k=5} & 56.64 &0.40 &36.68 &0.19 &30.91 &0.03 &28.38 &0.01 &34.54 &0.10 \\
\textit{w/ BGE Embedding, Top k=5} &67.29 &0.47 &43.39 &0.28 &28.31 &0.03 &32.22 &0.07 &36.95 &0.14 \\
\textit{w/ Openai Embedding, Top k=10} & 67.07 &0.53 &44.30 &0.27 &34.31 &0.05 &34.03 &0.06 &40.17 &0.15 \\
\textit{w/ BGE Embedding, Top k=10}& 71.74 &0.54 &47.68 &0.30 &30.55 &0.03 &30.57 &0.03 &38.80 &0.14 \\
\textit{w/ Openai Embedding, Top k=30}&66.27 &0.46 &46.28 &0.31 &38.95 &0.05 &46.15 &0.22 &45.42 &0.19 \\
\textit{w/ BGE Embedding, Top k=30}&57.35 &0.41 &46.92 &0.29 &35.30 &0.05 &42.82 &0.20 &42.04 &0.18 \\
\textit{w/ Openai Embedding, Top k=50}&55.94 &0.32 &47.94 &0.31 &34.32 &0.03 &46.64 &0.21 &42.60 &0.16 \\
\textit{w/ BGE Embedding, Top k=50}&59.41 &0.39 &38.52 &0.21 &35.40 &0.06 &45.47 &0.24 &41.51 &0.17 \\

\midrule
\multicolumn{11}{c}{\textbf{ $\mathtt{Set3}$ (100K-200K)}} \\
Qwen2-72B-Instruct (128K) & 46.99 & 0.27 & 37.06 & 0.13 & 31.50 & 0.02 & 35.01 & 0.07 & 35.94 & 0.09\\ 
\textit{w/ Openai Embedding, Top k=5} &63.91 &0.44 &33.56 &0.17 &25.98 &0.01 &28.98 &0.04 &34.48 &0.12 \\
\textit{w/ BGE Embedding, Top k=5} &64.81 &0.47 &30.27 &0.14 &25.88 &0.01 &27.86 &0.05 &33.70 &0.12 \\
\textit{w/ Openai Embedding, Top k=10} &67.50 &0.46 &33.44 &0.16 &27.94 &0.02 &31.62 &0.06 &36.47 &0.13 \\
\textit{w/ BGE Embedding, Top k=10}& 75.88 &0.56 &33.76 &0.15 &27.20 &0.01 &30.17 &0.04 &37.28 &0.14 \\
\textit{w/ Openai Embedding, Top k=30}&73.69 &0.55 &42.20 &0.27 &32.78 &0.02 &37.65 &0.13 &42.60 &0.18 \\
\textit{w/ BGE Embedding, Top k=30}&67.50 &0.47 &42.42 &0.18 &32.34 &0.03 &37.85 &0.12 &41.35 &0.15 \\
\textit{w/ Openai Embedding, Top k=50}&67.44 &0.50 &41.82 &0.24 &31.59 &0.04 &37.29 &0.12 &40.90 &0.18 \\
\textit{w/ BGE Embedding, Top k=50}&62.56 &0.42 &40.41 &0.18 &29.82 &0.02 &40.31 &0.14 &39.84 &0.15 \\

\midrule
\multicolumn{11}{c}{\textbf{$\mathtt{Set4}$ (200K-250K)}} \\
Qwen2-72B-Instruct (128K) & 33.18 & 0.16 & 26.59 & 0.08 & 29.84 & 0.01 & 25.81 & 0.04 & 28.92 & 0.06\\ 
\textit{w/ Openai Embedding, Top k=5} &51.49 &0.26 &17.12 &0.03 &21.59 &0.00 &16.37 &0.00 &25.59 &0.06 \\
\textit{w/ BGE Embedding, Top k=5} &48.40 &0.26 &14.55 &0.00 &20.69 &0.00 &18.07 &0.00 &24.63 &0.05 \\
\textit{w/ Openai Embedding, Top k=10} & 50.32 &0.28 &20.30 &0.03 &24.56 &0.00 &16.38 &0.00 &27.08 &0.06 \\
\textit{w/ BGE Embedding, Top k=10}& 51.02 &0.28 &21.88 &0.03 &25.45 &0.00 &17.29 &0.00 &28.10 &0.06 \\
\textit{w/ Openai Embedding, Top k=30}&52.17 &0.24 &24.60 &0.10 &26.78 &0.00 &17.79 &0.00 &29.29 &0.07 \\
\textit{w/ BGE Embedding, Top k=30}&47.98 &0.21 &26.82 &0.10 &26.70 &0.00 &20.02 &0.00 &29.44 &0.06 \\
\textit{w/ Openai Embedding, Top k=50}&51.63 &0.26 &23.62 &0.08 &24.49 &0.00 &21.84 &0.02 &29.14 &0.07 \\
\textit{w/ BGE Embedding, Top k=50}&47.23 &0.28 &27.78 &0.08 &26.48 &0.00 &24.44 &0.02 &30.52 &0.08 \\

\bottomrule
\end{tabular}
}
\caption{The result of adding RAG module on Qwen2-72B-Instruct with different length sets. }
\label{tb:rag_qwen2}
\end{table*}
\begin{table*}[t]
\centering
\begin{tabular}{p{0.96\textwidth}}
\toprule
\rowcolor{mygray}\multicolumn{1}{c}{\textbf{\textit{LongBench}}} \\
\midrule
\textbf{Passage 1:} The Real Glory. The Real Glory is a 1939 Samuel Goldwyn Productions adventure film starring \colorbox{myorange}{Gary Cooper}, David Niven, Andrea Leeds and Broderick Crawford released by United Artists in the weeks immediately following Nazi Germany's invasion of Poland. Based on a 1937 novel of the same name by \colorbox{myorange}{Charles L. Clifford} and directed by Henry Hathaway, the film is set against the backdrop of the Moro Rebellion during the American occupation of the Philippines at the beginning of the 20th century {\boldmath $\ldots$}\\
\textbf{Passage 2:} Jay Sheffield. Jay Howard Sheffield (September 25, 1934 – June 25, 1998) was an American actor, who appeared on the stage, in films, and on television. He married Barbara Babcock on June 9, 1962, in San Mateo, California. They later divorced {\boldmath $\ldots$} \\
\textbf{Passage 3:} David Niven. James David Graham Niven (; 1 March 1910 – 29 July 1983) was a British actor, soldier, memoirist, and novelist. Niven was known as a handsome and debonair leading man in Classic Hollywood films. He received an Academy Award and a Golden Globe Award {\boldmath $\ldots$}\\
\textbf{Passage 4:} Phileas Fogg snacks. Phileas Fogg Ltd is a company that produces snack products in the United Kingdom that was created in 1982 by Derwent Valley Foods. The brand is named for Phileas Fogg, the protagonist of Jules Verne's Around the World in Eighty Days {\boldmath $\ldots$}\\
\textbf{Passage 5:} Jules Verne Trophy. The Jules Verne Trophy is a prize for the fastest circumnavigation of the world by any type of yacht with no restrictions on the size of the crew provided the vessel has registered with the organization and paid an entry fee. A vessel holding the Jules Verne trophy will not necessarily hold the absolute round the world record {\boldmath $\ldots$}\\
\textbf{Question:} The actor that plays Phileas Fogg in "Around the World in 80 Days", co-starred with Gary Cooper in a 1939 Goldwyn Productions film based on a novel by what author?\\
\textbf{Answer:} Charles L. Clifford\\
\midrule
\rowcolor{mygray}\multicolumn{1}{c}{\textbf{\textit{Loong (Ours)}}} \\
\midrule
\textbf{Document 1:} \colorbox{myorange}{THE ARENA GROUP HOLDINGS, INC}. Proceeds from Simplify loan: \$7,748. Unearned revenue: \$(11,665). Amortization of debt discounts: \$536. \colorbox{myorange}{Cash and cash equivalents: \$4,003}. Noncash and accrued interest: \$2,839. Loss on impairment of assets: \$40,589. Accounts receivable, net: \$12,029. Subscription refund liability: \$18. Accounts payable: \$(102). Subscription acquisition costs: \$6,131. Change in fair value of contingent consideration: \$313. \colorbox{myorange}{(\$ in thousands, except share data)} {\boldmath $\ldots$} \\
\textbf{Document 2:} \colorbox{myorange}{General Enterprise Ventures, Inc}. For purposes of balance sheet presentation and reporting of cash flows, the Company considers all unrestricted demand deposits, money market funds and highly liquid debt instruments with an original maturity of less than 90 days to be cash and cash equivalents. The Company \colorbox{myorange}{did not have any cash equivalents} at March 31, 2024. The Company had cash of \colorbox{myorange}{\$549,755} at March 31, 2024 {\boldmath $\ldots$}\\
\textbf{Document 3:} \colorbox{myorange}{BROAD STREET REALTY, INC}. The carrying amounts of cash and cash equivalents, restricted cash, receivables and payables are reasonable estimates of their fair value as of March 31, 2024 due to the short-term nature of these instruments. Reconciliation of cash and cash equivalents and restricted cash:       Cash and cash equivalents: \colorbox{myorange}{\$14,631 (in thousands)} {\boldmath $\ldots$}\\
\textbf{Question:} Please list the `Cash and Cash Equivalents' of the aforementioned companies in ascending order. \\
\textbf{Answer:} 1. General Enterprise Ventures, Inc.: \$549,755. 2. Arena Group Holdings, Inc.: \$4,003 in thousands. 3. Broad Street Realty, Inc.: \$14,631 in thousands.\\
\bottomrule
\end{tabular}
\caption{Comparison of Evidence Distribution in Examples from LongBench and Loong (Ours). \textbf{Evidence} related to the answers is highlighted with \colorbox{myorange}{Orange Background}.}
\label{table:evidence}
\end{table*}

\end{document}